\documentclass{article}

\usepackage[nonatbib,preprint]{neurips_2025}

\usepackage[utf8]{inputenc} %
\usepackage[T1]{fontenc}  %
\usepackage{hyperref} %
\usepackage{url}  %
\usepackage{nicefrac} %

\usepackage{times,latexsym}
\usepackage{microtype}
\usepackage{subcaption}
\usepackage{makecell}
\usepackage[most, breakable, many]{tcolorbox}
\usepackage{amsmath}
\usepackage{amsthm}
\usepackage{amsfonts}
\usepackage{amssymb}
\usepackage{mathtools}
\usepackage{fancyvrb}
\usepackage{threeparttable}

\usepackage{tikz}
\usetikzlibrary{bayesnet}
\usetikzlibrary{arrows}
\usetikzlibrary{decorations}

\usepackage{graphicx}
\usepackage{booktabs}
\usepackage{tabularx}
\usepackage{tabulary}
\usepackage{colortbl}
\usepackage{xspace}
\usepackage{xcolor}
\newcolumntype{Y}{>{\centering\arraybackslash}X}
  \newcolumntype{P}{>{\raggedleft\arraybackslash}X}
\usepackage{multirow}
\usepackage{stfloats}
\usepackage{enumitem}

\usepackage{algorithm}
\usepackage{algpseudocode}
\usepackage{xpatch}
\usepackage{bm}

\usepackage[noabbrev,capitalise]{cleveref}

\newif\ifcommentsenabled
\commentsenabledfalse

\usepackage{amsmath,amsfonts,bm}

\def\eqref#1{equation~\ref{#1}}

\def\1{\bm{1}}

\DeclareMathAlphabet{\mathsfit}{\encodingdefault}{\sfdefault}{m}{sl}
\SetMathAlphabet{\mathsfit}{bold}{\encodingdefault}{\sfdefault}{bx}{n}

\usepackage{todonotes}
\newcommand*\iftodonotes{\if@todonotes@disabled\expandafter\@secondoftwo\else\expandafter\@firstoftwo\fi}
\makeatother

\definecolor{edolime}{rgb}{0.9,1,0.3}

\definecolor{obired}{rgb}{0.65,0.0,0.0}

\definecolor{omablue}{rgb}{0.04,0.8,0.95}

\newcommand{\acronym}{SEMI\xspace}

\title{Sample-efficient Integration of New Modalities \\ into Large Language Models}

\newcommand{\edin}{\epsilon}

\author{%
  Osman Batur İnce$^{\edin}$ \, \, André F. T. Martins$^{\tau, \lambda, \upsilon}$ \, \, Oisin Mac Aodha$^{\edin}$ \, \,  Edoardo M. Ponti$^{\edin}$\\
$^{\edin}$University of Edinburgh~~~~$^{\tau}$Instituto de Telecomunicações\\
$^{\lambda}$Instituto Superior Técnico, Universidade de Lisboa~~~~$^{\upsilon}$Unbabel\\
}

\begin{document}

\maketitle

\begin{abstract}
Multimodal foundation models can process several modalities. However, since the space of possible modalities is large and evolving over time, training a model from scratch to encompass all modalities is unfeasible. Moreover, integrating a modality into a pre-existing foundation model currently requires a significant amount of paired data, which is often not available for low-resource modalities. In this paper, we introduce a method for \textit{sample-efficient modality integration} (\acronym) into Large Language Models (LLMs). To this end, we devise a hypernetwork that can adapt a shared projector---placed between modality-specific encoders and an LLM decoder---to any modality. The hypernetwork, trained on high-resource modalities (i.e., text, speech, audio, video), is conditioned on a few samples from any arbitrary modality at inference time to generate a suitable adapter. To increase the diversity of training modalities, we artificially multiply the number of encoders through isometric transformations. We find that \acronym achieves a significant boost in sample efficiency during few-shot integration of new modalities (i.e., satellite images, astronomical images, inertial measurements, and molecules) with encoders of arbitrary embedding dimensionality. For instance, to reach the same accuracy as 32-shot \acronym, training the projector from scratch needs 64$\times$ more data. As a result, \acronym holds promise to extend the modality coverage of foundation models.\footnote{Our code and models are available at \href{https://github.com/ospanbatyr/sample-efficient-multimodality}{github.com/ospanbatyr/sample-efficient-multimodality}.}
\end{abstract}

\section{Introduction} \label{sec:intro}

Multimodal Foundation Models (MFMs) can perceive multiple modalities in input. 
Despite recent attempts to train ``omni-modal'' models \cite{unifiedio2, shukor2023unival, Qwen2.5-Omni}, these typically cover only a pre-defined and limited set of modalities.
As AI-based solutions are introduced into new fields and problem settings, the set of relevant modalities grows. 
As a consequence, it has become crucial to develop strategies to integrate new modalities incrementally into existing models \cite{Han2023OneLLMOF, Yu2024LLMsCE} without
na\"ively re-training them from scratch, which is extremely resource-intensive \cite{jiang2024specific}. A widely established practice consists of training a projector between each modality-specific encoder and a shared Large Language Model (LLM) decoder \cite{llama3} in a modular fashion \cite{pfeiffer2023modular}, thus recycling the pre-trained components. While being more compute-efficient than re-training from scratch, this often requires a large amount of paired data containing samples of the new modality and text. This crucial limitation makes integration unfeasible for low-resource modalities and burdensome for high-resource ones.

\begin{figure}[!ht]
  \centering
  \includegraphics[width=0.99\linewidth]{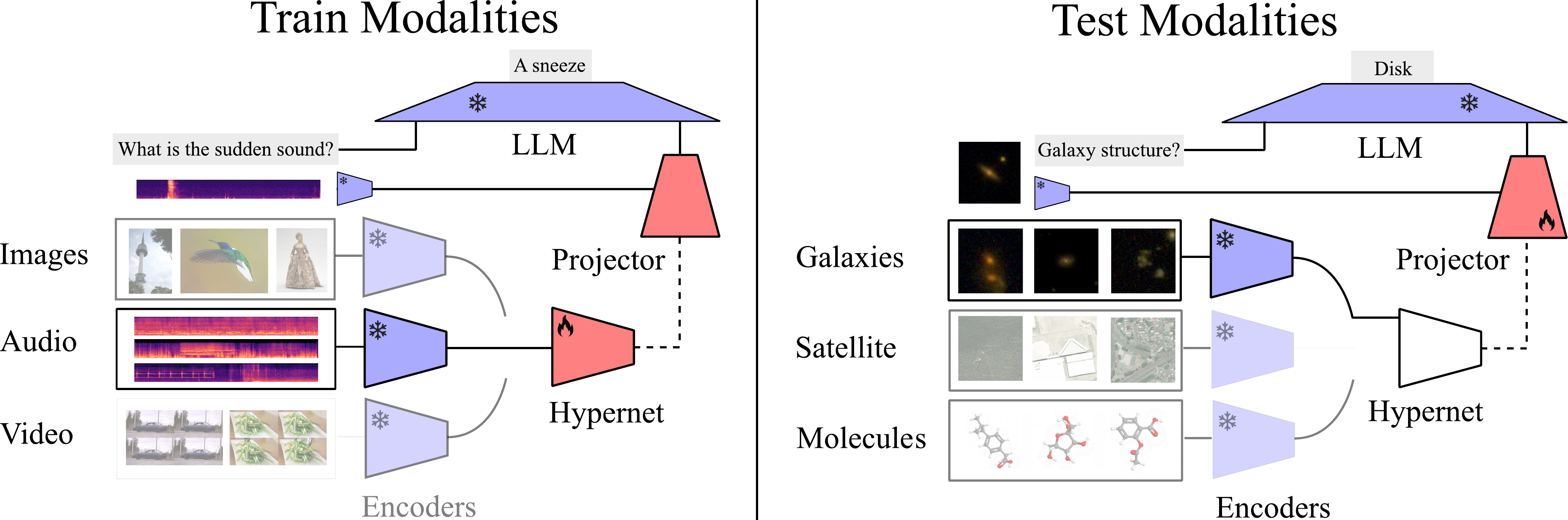}
  \caption{High-level framework of \acronym. \textbf{Left:} A hypernetwork is trained to generate an appropriate projector adaptation for high-resource training modalities. \textbf{Right:} A projector is generated by the hypernetwork for any unseen low-resource modality given only a few samples and is then fine-tuned on the same data. This enables the integration of new modalities with minimal training and paired data. Note that some modalities (audio, molecules) appear as images for visualisation purposes only.}
  \label{fig:high-level}
\vspace{-0.5em}
\end{figure}

To this end, we propose a novel paradigm called \textit{sample-efficient modality integration} (\acronym), which tackles the fundamental question: {how to integrate new modalities into foundation models given only a minimal set of samples?} Specifically, we partition modalities into two groups: a few high-resource training modalities---i.e., image, audio, and video---and low-resource test modalities---i.e., satellite images, galaxies, inertial measurement unit (IMU) data, and molecules---which we hold out to simulate new modalities that may emerge in real-world scenarios. We posit that learning the fundamental structure of modality integration from a subset of resource-rich modalities is sufficient to extrapolate to most other modalities (see \cref{fig:high-level}).

We introduce a three-stage paradigm for \acronym. In the first stage, we pre-train a non-linear projector between encoders of training modalities and an LLM decoder (both frozen) on the task of modality-to-text generation. In the second stage, we train a hypernetwork \cite{schmid1992hypernets,ha2017hypernetworks} on multimodal instruction data. The hypernetwork generates LoRAs \cite{hu2021loralowrankadaptationlarge}, conditioned on samples from training modalities, to adapt the shared projector from the first stage. 

In the third stage, we evaluate few-shot adaptation to 4 unseen (i.e., held-out during training) low-resource modalities, namely satellite images, astronomical images, sensor (IMU) data, and molecules. These represent a spectrum of modalities, from most similar to least similar to the training modalities, chosen to demonstrate the effectiveness of sample-efficient modality integration on a diverse set of real-world use cases. At the start of the third stage, the hypernetwork generates an adapter for each unseen modality given a small set of samples. The adapter is then fine-tuned on this very same data. By measuring how the model performance varies with increasing sample sizes, we find a significant boost in sample efficiency by virtue of our method. Specifically, we show that \acronym creates effective projectors with as few as 32 samples while other baselines fail, and often remains the best approach even in comparably larger-scale data regimes.

In summary, we offer the following main contributions: \textbf{1)} enabling the integration of low-resource modalities that have small-scale paired modality--text data but large-scale modality-only data; \textbf{2)} providing a systematic comparison of different baseline approaches for the newly defined challenge of sample-efficient modality integration; \textbf{3)} curating a collection of benchmarks for this challenge, including the creation of a new dataset for astronomical imaging;  \textbf{4}) proposing inexpensive strategies to augment the number of training `modalities' through isometric transformations of encoder outputs and generalise to arbitrary-dimensionality encoders. As a consequence, our work opens new opportunities to apply AI-based solutions to modalities that are resource-poor due to privacy constraints or the difficulty of collecting large quantities of paired data.

\section{Related Work} \label{sec:background}

\paragraph{Integrating a New Modality into Language Models}
Modalities different from text can be integrated into LLMs through several approaches, usually requiring large amounts of paired modality--text data. The most common one involves learning an MLP projector to map modality-specific encoder outputs onto the LLM input space \cite{mokady2021clipcapclipprefiximage, zhu2023minigpt, koh2023grounding, gao2023llamaadapterv2, liu2023llava}. Alternative methods include representing modality data as discrete tokens \cite{ge2023making}, incorporating trainable cross-attention layers within the LLM \cite{visualgpt}, employing more complex projectors like Q-Former \cite{blip2} or Perceiver \cite{perceiver, llava-uhd}, or combining cross-attention with a Perceiver resampler \cite{flamingo}. In the present work, we empirically demonstrate that the common strategy of na\"ively training an MLP projector from scratch for each new modality does not effectively integrate low-resource modalities into LLMs. 

\paragraph{Incremental Integration of Multiple Modalities}
MFMs often directly build upon the approaches listed above to integrate multiple modalities, thereby inheriting their data inefficiency. For instance, \cite{Zhao2023ChatBridgeBM, moon-etal-2024-anymal} train a separate projector for each modality, without any parameter sharing. Others improve upon this paradigm, 
by incrementally aligning modalities during training: this is achieved through a shared encoder paired with modality-specific tokens, which are then routed to projector experts in OneLLM
\cite{Han2023OneLLMOF}, or a combination of uni-modal and cross-modal adapters in PathWeave \cite{Yu2024LLMsCE}. 
We compare against baselines inspired by these incremental modality adaptation strategies, where we adapt shared projectors with parameter-efficient fine-tuning. Contrary to the setup in OneLLM \cite{Han2023OneLLMOF}, we assume modality-specific encoders to be given, instead of jointly trained. This modularity adds more flexibility and is compatible with integrating existing off-the-shelf encoders.

On the other hand, other strategies for integrating multiple modalities do not meet the desiderata for our setup, namely, sample efficiency and compatibility with generative text models.
Incorporating new modalities by combining existing MFMs \cite{chen-etal-2024-model} is unrealistic for our setup, as it assumes that MFMs for a low-resource modality exist in the first place. While \cite{liu2024towards} focused on ``\textit{enabling models to generalise to unseen modalities}'', it did not demonstrate how to integrate new modalities {into generative LLMs}. Finally, other strategies---such as EE-MLLM \cite{eemllm}, Macaw-LLM \cite{macawllm}, and ImageBind-LLM \cite{imagebindllm}---either insufficiently substantiate their sample efficiency or implicitly assume the availability of large-scale paired data for low-resource modalities during prior training stages. This prevents them from serving as baselines for our low-resource integration challenge.

\section{Sample-Efficient Modality Integration} \label{sec:method}

In this work, we pragmatically define any two `modalities' to be distinct if they correspond to different distributions of encoder outputs. However, a finer distinction can also be made between a new encoder for the same data distribution, a change in probability function $p$ for the same sample space $\Omega$ (e.g., a domain shift), and a change in both $p$ and $\Omega$, for instance, RGB images $\{0,\dots,255\}^{H\times W\times3}$ versus directed graphs $\{0, 1\}^{N\times N}$.

Our main goal is to devise an efficient and effective way of integrating low-resource data modalities into LLMs.
We aim to develop a general framework that makes minimal assumptions about the distributions of data for unseen modalities or the encoder architectures they use. Although several approaches with these properties exist (see \cref{sec:background}), we start from the most widespread and conceptually simple approach that only requires an MLP projector between the modality encoders and the LLM decoder.\footnote{While our framework is compatible with other approaches such as cross-attention or Q-Former, we leave these extensions to future work.}

For \acronym, we propose a solution consisting of three stages: \textbf{1}) we train a shared MLP projector with coarse-grained data from resource-rich modalities (\cref{ssec:phase1}); \textbf{2}) we then train a hypernetwork \cite{schmid1992hypernets, ha2017hypernetworks} to generate projector adapters with instruction data from resource-rich modalities (\cref{ssec:phase2}). Intuitively, this aims to transfer the ability to adapt to a modality from high-resource modalities to low-resource ones; \textbf{3}) finally, we fine-tune the adapter generated for a new, resource-poor modality with only a few data points (\cref{ssec:phase3}). Additionally, we show how \acronym can generalise to new encoders of arbitrary dimensionality (\cref{ssec:arbitrary_dimensions}). We present an overview of this process in \cref{fig:high-level}.

\subsection{Training the Shared Projector on Train Modalities}
\label{ssec:phase1}
Assume we are given $M$ encoders $\{\text{enc}_m\}_{m=1}^M$ for high-resource modalities and a decoder LLM, whose parameters are all frozen. First, we train a projector $\text{proj}_\psi(\cdot) : \mathbb{R}^{h_e} \rightarrow \mathbb{R}^{h_d}$ on paired raw data from these modalities (i.e., image--text, video--text, and audio--text), where $h_e$ and $h_d$ are the encoder output and decoder input dimensions, respectively. This establishes a universal mapping between observed encoders and the LLM decoder.

\subsection{Training the Hypernetwork on Train Modalities}
\label{ssec:phase2}
Afterwards, we train a hypernetwork to generate modality-specific adapters to be composed with the shared projector given only a small data sample, effectively simulating the desired test-time few-shot adaptation to low-resource modalities. The first step in hypernetwork training is sampling a training modality encoder $\text{enc}_m$, an instruction $\mathbf{i}_m \in \mathcal{I}_{m}$ from the instruction pool belonging to that modality, and a sample of examples $\{\mathbf{x}_m, \mathbf{y}_m\}_1^S \in \mathcal{D}_m$: each example consists in a modality-specific input $\mathbf{x}_m$ and text $\mathbf{y}_m$ to better ground the modality. The hypernetwork receives the corresponding interleaved encodings $(\text{enc}_\text{text}(\mathbf{i}_m) \oplus [\text{enc}_m(\mathbf{x}_m) \oplus \text{enc}_\text{text}(\mathbf{y}_m)]_1^S)$ and generates LoRA adapters $\delta$. 

After sampling a separate batch $\{\mathbf{x}_m, \mathbf{y}_m\}_1^B$, the LoRA adapter is plugged into the parameters of the shared projector (now frozen) to project the encoding of the modality-specific input into the LLM token space. This yields $\text{proj}_{\psi + \delta}(\mathbf{x}_m)$, which, combined with the instruction $\mathbf{i}$, is fed into the LLM decoder. 
The final loss is the cross-entropy of the LLM decoder prediction with respect to the target texts $\{\mathbf{y}_m\}_1^B$, whose gradient is back-propagated to the hypernetwork parameters for optimisation.
Note that the sample size for the hypernetwork $S$ and the batch size of the LLM decoder $B$ may be different, and thus can be chosen arbitrarily. Nevertheless, $S$ is preferably chosen to be small due to the limited amount of text-paired data one might collect for test modalities. Additionally, we demonstrate that larger context lengths do not necessarily improve performance (see \cref{tab:ablations}).

We employ several techniques to enhance performance and simplify the optimisation process. These techniques include factorising the hyper-network's generated parameters, using isometric transformations to emulate numerous encoders, and grounding modality inputs with text when feeding them to the hypernetwork, as expounded in the following paragraphs. The pseudocode for the shared projector training, as well as hypernetwork training and adaptation, is provided in \cref{sec:pseudocodes}.

\textbf{Hypernetwork Optimisation}~~~Hypernetworks often present optimisation challenges and high computational complexity. For instance, predicting all parameters of an $N \times M$ weight matrix from a $K$-dimensional embedding requires $N \times M \times K$ parameters in the hypernetwork's generating linear layer. In our setup, we reduce complexity by generating lower-rank adapter parameters $\delta$ representing the difference between the pre-trained projector $\text{proj}_\psi$ and the target modality projector $\text{proj}_{\psi + \delta}$ instead of the full projector weights. This change requires $(N + M)\times R \times K$ parameters, and depending on the LoRA rank $R \ll M$, it can remediate the parametric complexity significantly. This also alleviates optimisation challenges, particularly initialisation issues \cite{principledweightinit, hypnetsurvey}.

\textbf{Emulating Many Encoders}~~~Ideally, our hypernetwork should infer the statistical properties and characteristics of a new modality $m^\prime$ through the lens of a modality encoder $\text{enc}_{m^\prime}$ given only a small number of samples. Therefore, training the hypernetwork on numerous modality encoders so that it generalises better, rather than over-fitting to a few encoders, is preferred. However, if we simply scaled the number of modalities by sourcing more readily available, pre-trained encoders, we would soon encounter a barrier due to their scarcity.

Hence, we use random orthogonal matrices to emulate new encoders, sampling them from an $O(d_h)$ Haar distribution \cite{haar, scipy}, where $d_{h}$ denotes the hypernetwork dimension, and we transform modality-specific encodings through these matrices before feeding them as input to the hypernetwork or the adapted projector. Orthogonal matrices possess desirable properties like invertibility and isometry, preserving the Euclidean distance and the inner product of vectors. For instance, orthogonal matrices encompass rotation, reflection, and permutation, among other types of linear transformations. By using these random transformations, we can emulate numerous encoders by altering the general distribution of the data while preserving certain characteristics along with the local spatial relationships between instances.

\textbf{Grounding with Text}~~~Increasingly more bimodal MFMs incorporate text conditioning into modality projections \cite{instructblip, kar2024brave, emma}. As our hypernetwork must generalise across diverse unseen modalities and encoder distributions, the same embedding may hold different meanings in different encoder spaces. This is because each modality encoder $\text{enc}_m$ learns a unique mapping from the input data to its embedding space, influenced by its architecture and training data, among other factors. If the hypernetwork treats these embeddings as universally comparable without taking this variation into account, it can generate inaccurate adapters for new modalities. To overcome this issue, we provide the hypernetwork with both instruction embeddings $\text{enc}_\text{text}(\mathbf{i}_m)$ and text embeddings $\text{enc}_\text{text}(\mathbf{y}_m)$ alongside modality embeddings $\text{enc}_m(\mathbf{x}_m)$. Note that instruction and text embeddings are extracted from the same frozen text encoder $\text{enc}_\text{text}$ throughout training and inference, effectively grounding modality embeddings on a fixed representation space.  

\subsection{Adaptation of the Generated Adapters for Test Modalities}
\label{ssec:phase3}
Finally, during adaptation to a new test modality, the hypernetwork (as well as the LLM decoder and modality encoders) remains frozen. We partition the training data into batches with a maximum sequence length determined by the context length of the hypernetwork, and generate adapters for each batch. For each batch, similar to \cref{ssec:phase2}, an instruction and interleaved modality--text data are encoded and then fed to the hypernetwork, which generates adapters for the new modality. Adapters generated across batches are averaged ($\bar{\delta}$), then merged with the pre-trained projector parameters $\psi$ to create an updated projector $\text{proj}_{\psi + \bar\delta}$. We provide additional details on adapter generation techniques in \cref{sec:adapter_generation}. Finally, the merged projector is fine-tuned on the low-resource modality's few-shot samples, based on the cross-entropy loss of true and predicted output text. We illustrate the effects of this third stage on the representation similarity of text and each unseen modality in \cref{app:xmod_sim}.

\subsection{Integrating Arbitrary Dimensionality Encoders}
\label{ssec:arbitrary_dimensions}
Although we enforce a fixed hypernetwork input dimension, we demonstrate that \acronym can effectively generalise to encoders with varying output dimensions, a capability previously unexplored in this context to the best of our knowledge. To handle smaller encoder dimensions, we prune the pre-trained projector by removing the final dimensions from its weights and biases to match the encoder output dimension. To handle larger encoder dimensions, we utilise an efficient unsupervised feature selection method, Infinite Feature Selection (Inf-FS) \cite{infinitefs}, to reduce the encoder output dimensions. 
We hypothesise that Inf-FS is superior to alternative techniques such as PCA in settings (such as ours) where the number of training samples is much smaller than the number of dimensions. With $N$ samples, $d_{e}$-dimensional encoder, and $d_{h}$-dimensional hypernetwork, where $d_{h} < d_{e} \text{, }d_{e} \gg N \text{, and }d_{h} \gg N$, PCA's resulting embedding rank is limited to $N$. In contrast, as Inf-FS selects $d_{h}$ features, it can construct $d_{h}$-rank embeddings upper bounded by the rank of the original embeddings. We verify that Inf-FS is more stable than PCA with an ablation in \cref{app:dim_red}.

\section{Experimental Setup} \label{sec:exp_setup}
\textbf{Model Architecture} The pre-trained projector is a 2-layer MLP connecting modality-specific encoders to the LLM's input space. Our hypernetwork adopts an architecture where special tokens (one per generated layer) are concatenated with instruction--text--modality samples, whose encodings are combined with sinusoidal positional embeddings. An attention layer then contextualises the special token embeddings with respect to the samples. Finally, linear layers applied to each special token generate the corresponding adapter layers. In our setup, we generate adapters only for the first projector layer, leaving the second layer unchanged (see \cref{sec:arch_choice} for more details). We adopt Llama 3.1 8B Instruct \cite{llama3} and Llama 3.2 1B Instruct \cite{llama3.2_cards} as LLM decoders for our experiments. We report 8B LLM results in the main paper and 1B LM results in \cref{sec:add_metrics_1b}. We conducted ablations and other exploratory experiments with the 1B LLM unless otherwise specified. We use GTE-ModernBERT-Base \cite{zhang2024mgte, modernbert} as our text encoder. Details on compute resources, runtime, and hyperparameters are provided in \cref{sec:train_details}.

\begin{table}[!t]
\centering
\caption{Training modalities, datasets, and encoders. We generally use captioning datasets during the pre-training of the projector and instruction datasets during the hypernetwork training. The size of each dataset is shown in parentheses next to its name.}\label{tab:train_dset1}
\resizebox{\textwidth}{!}{\begin{tabular}{@{}lllll@{}}
\toprule
\multirow{2}{*}{\textbf{Modality}} & \multicolumn{2}{l}{\textbf{Stage 1 (Projector pre-training)}}  & \multicolumn{2}{l}{\textbf{Stage 2 (Hypernetwork training)}}   \\ \cmidrule(l){2-5} 
   & \textbf{Dataset}   & \textbf{Encoder}  & \textbf{Dataset}  & \textbf{Encoder}   \\ \midrule
\textbf{Text \& Image} & COCO \cite{lin2014microsoft} (590K) & CLIP \cite{radford2021clip}  & ShareGPT4V \cite{chen2023sharegpt4v} (35K) & SigLIP 2 \cite{siglip2}  \\
\textbf{Text \& Audio} & AudioCaps \cite{kim2019audiocaps} (45K) & CLAP \cite{elizalde2023clap} & Clotho-Detail \cite{bubogpt, clotho} (3.9K) & Cacophony \cite{cacophony}  \\
\textbf{Text \& Video} & OpenVid \cite{nan2024openvid} (59K) & VideoCLIP-XL \cite{wang-etal-2024-videoclip} & ShareGPT4Video \cite{chen2024sharegpt4video} (39K) & ViCLIP \cite{wang2024internvid} \\ \bottomrule
\end{tabular}}
\end{table}

\begin{table}[!ht]
\centering
\caption{Evaluation modalities with their datasets and encoders. All datasets are captioning datasets. We distinguish among modalities based on whether they are seen, constitute an unseen distribution of a seen input space (i.e., a domain shift), or are completely unseen during training (see \cref{sec:method}).}
\label{tab:test_dsets}

\resizebox{\textwidth}{!}{
\renewcommand{\arraystretch}{1.1}
\begin{tabular}{l|l|ll|ll}
\toprule
\textbf{Modality} & \textbf{Seen} & \multicolumn{2}{c}{\textbf{Unseen Domain}} & \multicolumn{2}{|c}{\textbf{Unseen Modality}}  \\
 & {Audio} & {Satellite Images} & {Astronomical Images} &  {IMU} & {Molecule} \\ 
\midrule
\textbf{Dataset} & SoundBible \cite{wavcaps} & SydneyCaptions \cite{sydneycaps} & CAPDELS (ours) & SensorCaps \cite{sensorcaps} & ChEBI-20 \cite{chebi20} \\ 
\textbf{Split Sizes} & 62 / 184 / 186 & 2485 / 290 / 290 & 4344 / 480 / 1311 & 1670 / 209 / 209 & 26407 / 3301 / 3300 \\ \vspace{0.25em}
\textbf{\begin{tabular}[l]{@{}l@{}}Encoder\\ Variants\\ Emb. Dim\end{tabular}} & {\begin{tabular}[l]{@{}l@{}}BLAT \cite{blat} \\ - \\ 768 \end{tabular}} & \multicolumn{1}{l}{\begin{tabular}[l]{@{}l@{}}RemoteCLIP \cite{remoteclip}\\ B-32 / L-14 / RN-50\\ 512 / 768 / 1024\end{tabular}} & \multicolumn{1}{l}{\begin{tabular}[l]{@{}l@{}}Zoobot ConvNeXt \cite{zoobot} \\ Nano / Tiny / Base\\ 640 / 768 / 1024\end{tabular}} & \multicolumn{1}{|l}{\begin{tabular}[l]{@{}l@{}}LIMU-BERT \cite{limubert} \\ - \\ 720\end{tabular}} & \multicolumn{1}{l}{\begin{tabular}[l]{@{}l@{}}MolCA \cite{molca} \\ - \\ 768\end{tabular}} \vspace{-0.25em}\\ \bottomrule
\end{tabular}}
\vspace{-1.5em}
\end{table}

\textbf{Modalities: Datasets and Encoders}~~~We trained our hypernetwork on image, audio, and video data, then evaluated its few-shot adaptation capabilities across a spectrum of shifts from the training encoders: a new encoder for a seen modality (audio), two unseen domains for images (satellite and galaxies), and two entirely unseen modalities (IMU data and molecules). This allowed us to assess how the performance of \acronym changes on modalities from most to least similar to the training data in a systematic way.
The chosen test modalities thus span a range of adaptation difficulties and diverse applied use cases of AI (in geolocation, astronomy, navigation, and biology/medicine). 

Table~\ref{tab:train_dset1} lists the modalities, datasets, and encoders employed during training. We followed the LLaVA framework \cite{liu2023llava} and used coarse captioning datasets for projector pre-training (except for video data) and fine-grained description datasets for hypernetwork training. To mitigate potential overfitting, we used different encoders for the same modality during the two stages. This mimics our intended new modality adaptation scenario, in which the encoders for new modalities are unseen.

As for test modalities, their datasets, encoders, and additional information are detailed in Table~\ref{tab:test_dsets}. 
In addition to sourcing existing datasets, we also created CAPDELS, a pioneering novel astronomical imaging captioning dataset (see Appendix \ref{sec:capdels} for details), built on the Galaxy Zoo CANDELS multi-label galaxy morphological classification dataset \cite{candels}. For both satellite and astronomical imaging, we used a family of encoders varying in size and embedding dimension---  
RemoteCLIP \cite{remoteclip} and Zoobot ConvNeXt \cite{zoobot}, respectively---but trained on the same dataset. This allowed us to make justifiable claims about the ability of \acronym to generalise to different encoder sizes. Note that the input spaces (images) for these two modalities are observed during training, while their distribution is shifted with respect to the domains of training images in COCO. In addition, we explore two entirely new input spaces: three-axis accelerometer and gyroscope numerical readings for IMU data and labelled graphs for molecules. For IMU data, after extracting IMU embeddings via LIMU-BERT, we performed dimensionality reduction by averaging groups of consecutive tokens, effectively reducing the temporal resolution while preserving the feature space. The resulting embeddings are flattened into a single vector.
Finally, we evaluated \acronym also on few-shot adaptation to an unseen encoder for one of the high-resource training modalities, namely audio. In this case, both the data domain and input space are similar to one of the training datasets (AudioCaps). This helps demonstrate the broad scope of our methods, which may benefit the integration of new encoders for seen, high-resource modalities, too. Additional dataset details, including pre-processing steps, are provided in \cref{sec:data_details}.

\textbf{Baselines}~~~To evaluate the impact of cross-modality transfer with our hypernetwork, we compared \acronym against three baselines that are representative of current state-of-the-art approaches. Our simplest baseline (\textit{Projector}) consists in training a (randomly initialised) projector from scratch on the few-shot examples of each low-resource modality. A second baseline (\textit{LoRA}) trains a LoRA adapter on few-shot examples and merges it with the pre-trained shared projector. Finally, \textit{FT Projector} fully fine-tunes the pre-trained shared projector on each test modality, representing our strongest baseline. \textit{LoRA} is reminiscent of PathWeave \cite{Yu2024LLMsCE} and \textit{FT Projector} of the OneLLM \cite{Han2023OneLLMOF} framework; however, to the best of our knowledge, these baselines constitute the first attempt to streamline these setups and make them comparable in a controlled setting.
As with our hypernetwork-based \acronym, we applied weight pruning or Inf-FS dimensionality reduction to \textit{LoRA} and \textit{FT Projector} when adapting them to smaller and larger encoder dimensionalities, respectively. Instead, the \textit{Projector} baseline was directly created with the target dimensions as it does not rely on the pre-trained projector.

\textbf{Evaluation}~~~We evaluate our method and the baselines using greedy decoding on the test set and calculating n-gram based metrics (BLEU-4 (BLEU) \cite{bleu}, METEOR \cite{meteor}, ROUGE-1, and ROUGE-2 \cite{rouge}), a longest common subsequence based metric (ROUGE-L \cite{rouge}), and the CIDEr \cite{cider} metric. Since CIDEr is designed for image description tasks, we excluded it from our IMU and molecule evaluation. We perform model selection through early stopping according to the model's CIDEr (or BLEU when unavailable) on the validation sets. 

Given our focus on sample-efficient modality integration, we evaluated \acronym against the baselines using a range of dataset sizes for few-shot adaptation. To ensure a fair comparison, we randomly selected subsets of varying sizes from each unseen modality dataset and tested all methods on these identical splits. Specifically, the subsets range from 32 samples up to the full dataset size, increasing by a factor of four (e.g., 32, 128, 512, 2048, and 2485 for the SydneyCaptions dataset). We trained and evaluated each method with the same three random seeds to ensure identical training batches (with the exception of the ChEBI-20 dataset, where a single seed was used due to its large size).

\section{Results} \label{sec:results}

\subsection{Main Results}

In this section, we report CIDEr scores for satellite and galaxy modalities, and BLEU scores for IMU, molecule, and audio modalities; additional metrics and qualitative examples are available in Appendices \ref{sec:add_metrics} and \ref{sec:qual_examples}, respectively. To measure sample efficiency, we study how these metrics vary as a function of the sample size for each new modality.

\paragraph{Satellite Images}

\begin{figure}[!t]
  \centering
  \includegraphics[width=\linewidth]{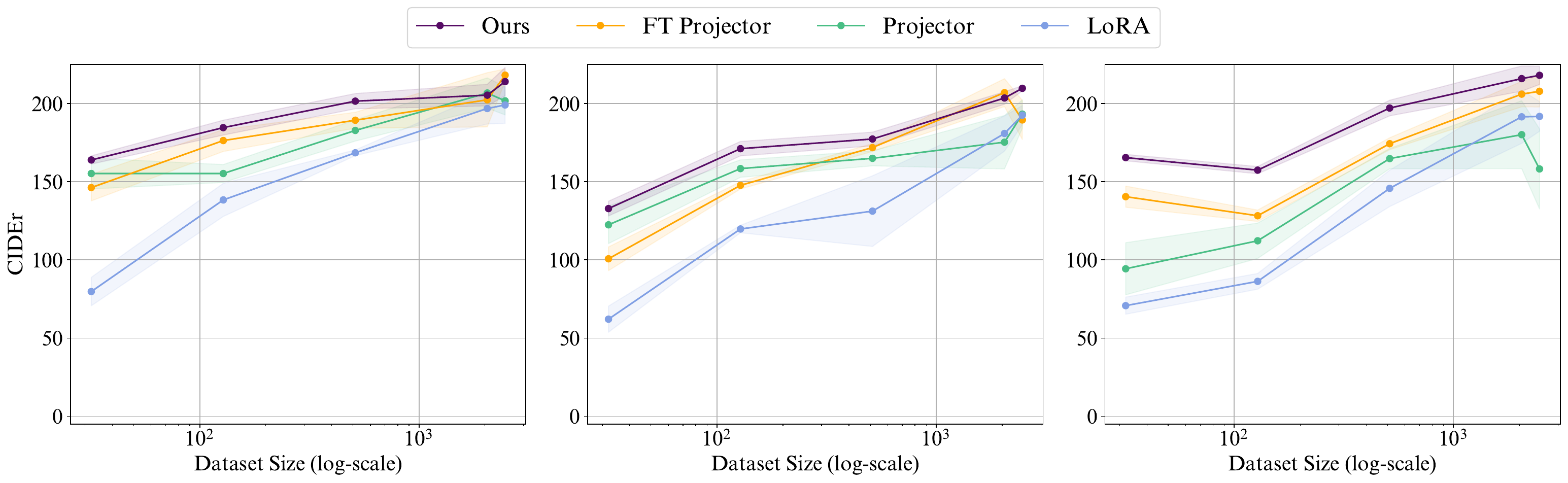}
  \hspace{1em}
  \begin{minipage}[t]{.31\linewidth}
  \vspace{-1em}
  \centering
  \subcaption{ViT-Base-32}\label{fig:sydney_vitb32}
  \end{minipage}%
  \begin{minipage}[t]{.31\linewidth}
  \vspace{-1em}
  \centering
  \subcaption{ViT-Large-14}\label{fig:sydney_vitl14}
  \end{minipage}%
  \hspace{1em}
  \begin{minipage}[t]{.31\linewidth}
  \vspace{-1em}
  \centering
  \subcaption{ResNet-50}\label{fig:sydney_rn50}
  \end{minipage}
  \caption{SydneyCaptions satellite captioning results with three different encoders. The shaded areas around the lines indicate the standard error obtained from multiple seeds.
  }
  \vspace{-1em}
  \label{fig:sydney}
\end{figure}

\begin{figure}[!t]
  \centering
  \includegraphics[width=\linewidth]{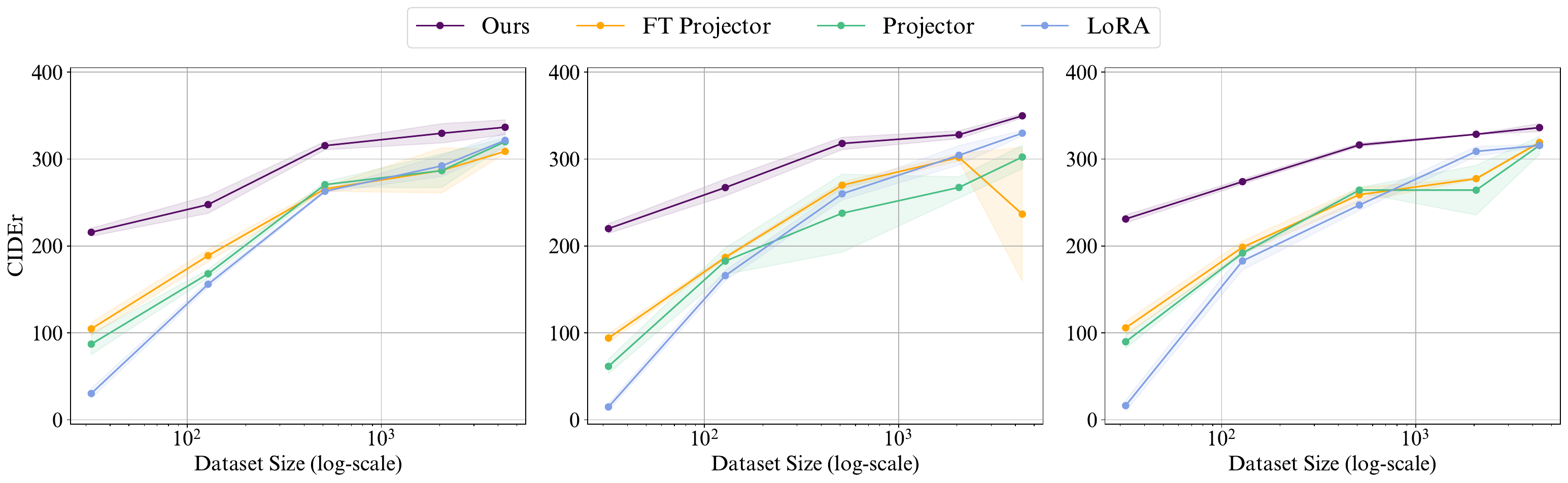}
  \begin{minipage}[t]{.32\linewidth}
  \vspace{-1em}
  \centering
  \subcaption{ConvNeXt-Nano}\label{fig:capdels_nano}
  \end{minipage}%
  \begin{minipage}[t]{.32\linewidth}
  \vspace{-1em}
  \centering
  \subcaption{ConvNeXt-Tiny}\label{fig:capdels_tiny}
  \end{minipage}%
  \begin{minipage}[t]{.32\linewidth}
  \vspace{-1em}
  \centering
  \subcaption{ConvNeXt-Base}\label{fig:capdels_base}
  \end{minipage}
  \caption{CAPDELS astronomical image captioning results with three different encoders. The shaded areas around the lines indicate the standard error obtained from multiple seeds.
  }
  \vspace{-1em}
  \label{fig:capdels}
\end{figure}

The CIDEr scores for the satellite imaging modality are shown in Figure~\ref{fig:sydney}. Overall, our approach outperforms (or at worst matches) all baselines across all sample sizes and encoder dimensionalities. The methods generally rank in descending order of performance as follows: \acronym, \textit{FT Projector}, \textit{Projector}, and \textit{LoRA}. However, note that \textit{LoRA} partially bridges the gap in the higher data regimes. In fact, as the sample size grows, the impact of the inductive bias provided by each method diminishes.

When comparing encoders of varying sizes and dimensionality, our method consistently performed well, effectively generalising to encoders with a smaller or larger embedding dimension than what was observed during training (see Table~\ref{tab:train_dset1} for the embedding dimensions of encoders). Importantly, our method exhibits the largest gains over baselines when integrating the encoder with the largest dimensionality, i.e. the ResNet-50 variant (Figure \ref{fig:sydney_vitl14}). These results highlight the positive scaling behaviour of \acronym.
As an additional finding, we observed that training projectors from scratch (\textit{Projector} baseline) with smaller-dimensionality encoders (ViT-Base-32) improved performance in low-data regimes over larger encoders (ResNet-50) by preventing overfitting. %

\paragraph{Astronomical Images} Although overall astronomical imaging results are similar to satellite imaging results, we note certain differences (Figure~\ref{fig:capdels}). In particular, the gap between \acronym and all baseline is significantly larger, especially in low-size and mid-size encoders, reaching a difference of 200 CIDEr points for ConvNeXt-Tiny and ConvNeXt-Base in 32-shot settings. On the other hand, the baselines perform comparably,
and they all exhibit remarkable variance, showcasing their brittleness compared with \acronym.

\begin{figure}[t]
  \centering
  \begin{minipage}{0.95\textwidth}
  \centering
  \includegraphics[width=.55\linewidth]{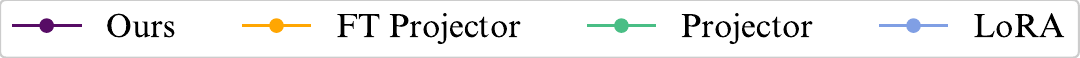}
  \end{minipage}

  \begin{minipage}{0.45\textwidth}
  \includegraphics[width=.95\linewidth]{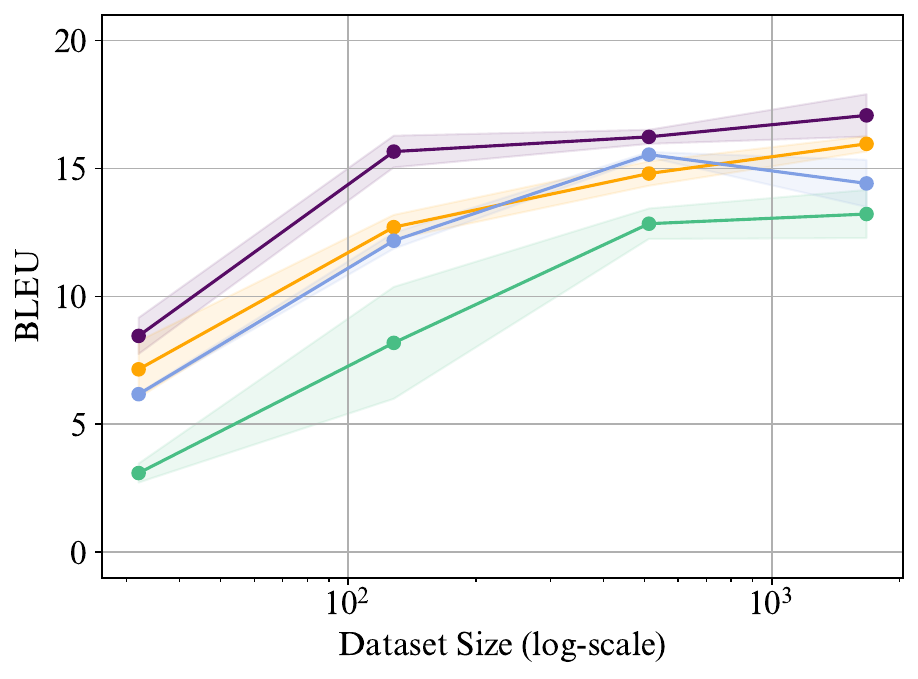}
  \end{minipage}
  \hfill
  \begin{minipage}{0.45\textwidth}
  \includegraphics[width=.95\linewidth]{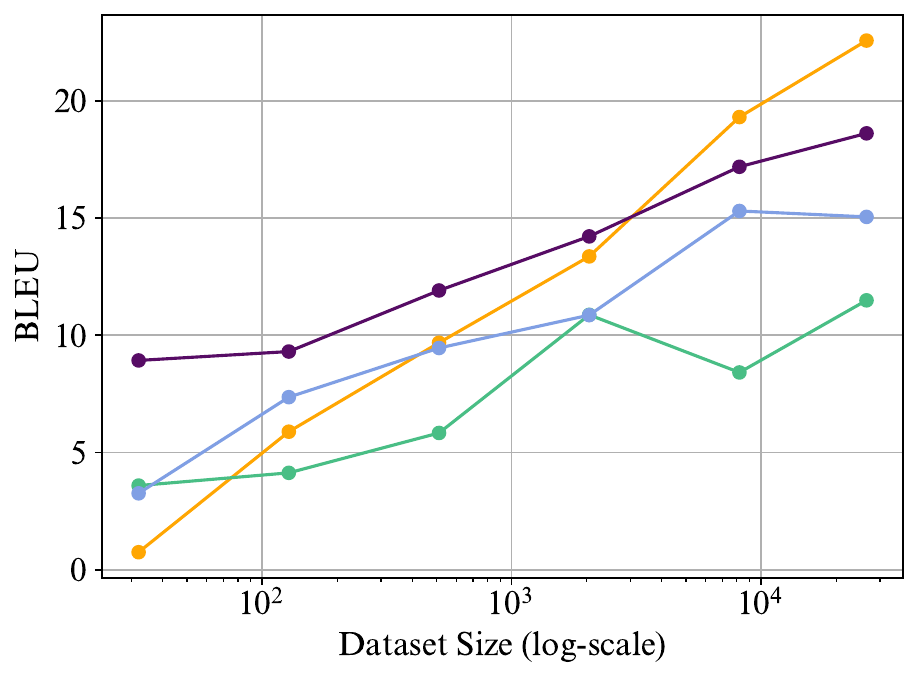}
  \end{minipage}
  \caption{\textbf{Left}: SensorCaps activity description dataset results. \textbf{Right}: ChEBI-20 molecule description dataset results. 
  }
  \vspace{-0.5em}
  \label{fig:sensorcaps_n_chebi20}
\end{figure}

\paragraph{IMU Data}
Focusing now on entirely novel modalities, \acronym outperforms all baselines across all sample sizes for IMU data (see \cref{fig:sensorcaps_n_chebi20} left). After our method, \textit{FT Projector} and \textit{LoRA} remain the second-best-performing methods, followed by the significantly weaker \textit{Projector}. This highlights the positive contribution of cross-modal transfer even when the distance between train and test modalities increases.
For instance, the best baseline, \textit{FT Projector}, requires 16 times more examples (2048) to achieve comparable performance to 128-shot \acronym.

\paragraph{Molecules}

Overall, we observe that our method outperforms all baselines (see \cref{fig:sensorcaps_n_chebi20} right) for molecules; however, differently from previous modalities, \textit{FT Projector} eventually catches up and surpasses \acronym at high-resource settings ($10^4$ examples). %
On the other hand, \acronym demonstrates particularly strong performance in low-resource settings. The best-performing baseline, \textit{FT Projector}, requires 16 times more data (512 samples) to reach comparable results to 32-shot \acronym. 

\paragraph{Audio}

\begin{figure}[t]
  \centering
  \begin{minipage}{0.95\textwidth}
  \centering
  \includegraphics[width=.55\linewidth]{figures/legend_8b.pdf}
  \end{minipage}

  \begin{minipage}{0.45\textwidth}
  \includegraphics[width=.95\linewidth]{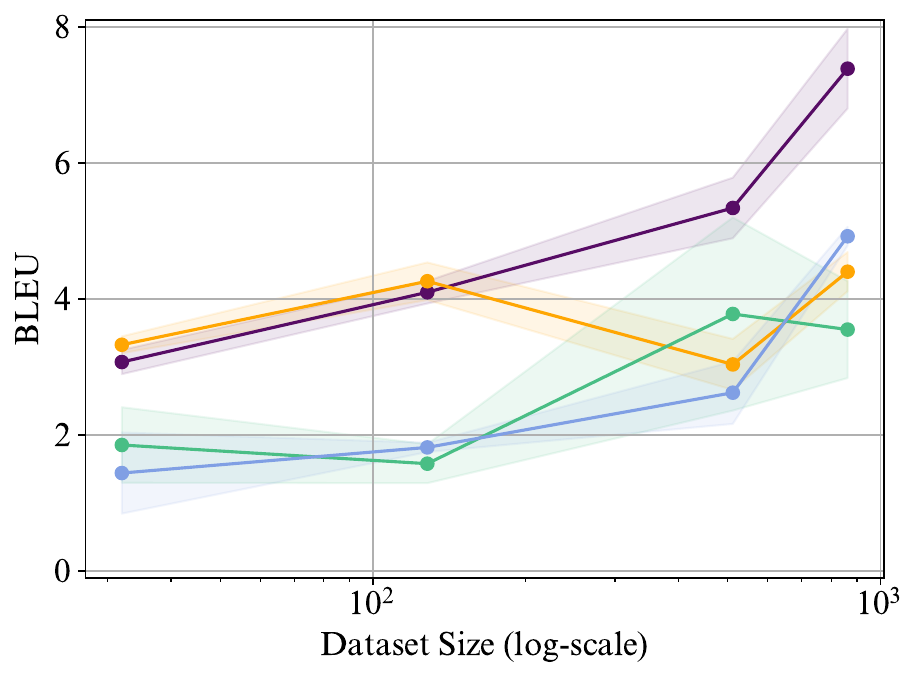}
  \end{minipage}
  \hfill
  \begin{minipage}{0.45\textwidth}
  \includegraphics[width=.95\linewidth]{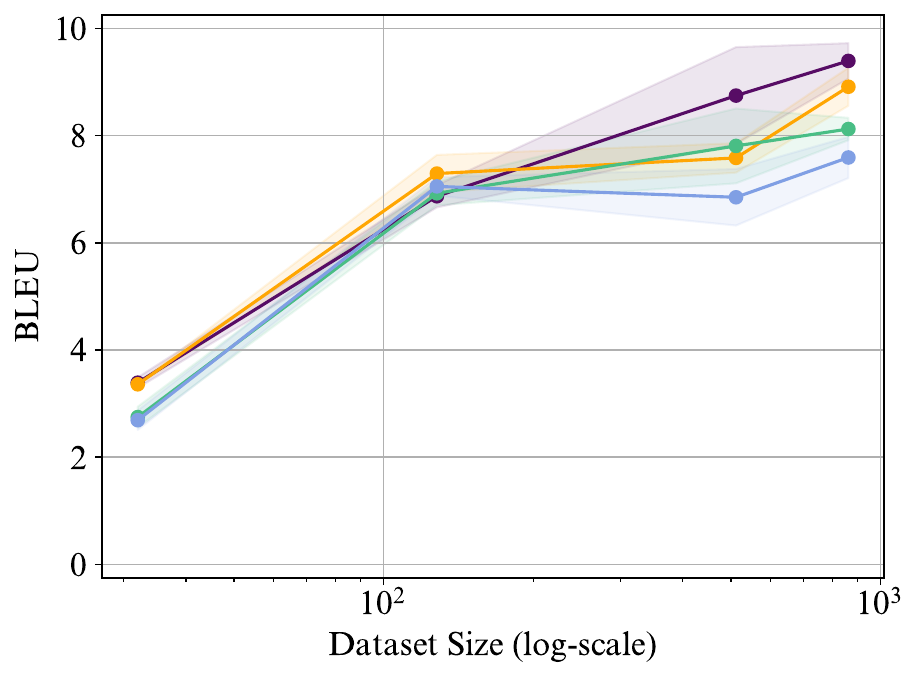}
  \end{minipage}
  \caption{SoundBible audio captioning results. \textbf{Left:} Llama 3.2 1B Instruct. \textbf{Right:} Llama 3.1 8B Instruct. 
  }
  \vspace{-0.5em}
  \label{fig:soundbible}
\end{figure}

To present a more comprehensive view of our approach, we also evaluated our method on a new encoder for a seen modality, audio captioning. The results are available in \cref{fig:soundbible}, where we additionally compare two LLM sizes (1B and 8B).
For the smaller LM (left) and the larger LLM (right), we observe that our approach and \textit{FT Projector} are comparable in extremely low data regimes ($<10^2$); however, the performance discrepancy between \acronym and the baselines widens as sample size increases (more significantly so in the smaller LLM).
This surprising finding suggests that hypernetwork-based \acronym may bring benefits to the integration of seen modalities, too.

\subsection{Ablations}

Finally, to justify the architecture of our hypernetwork, we conduct a series of ablations to demonstrate the impact of each of our design choices.
Table \ref{tab:ablations} in \cref{app:hyp_abl} shows that the combination of text grounding and isometric transformations results in the most accurate modality integration overall, especially for the CAPDELS dataset.
This stems from grounding all other modalities on text as an `anchor' modality and from avoiding overfitting by artificially multiplying the number of encoders, respectively.
On the other hand, a 2-layer transformer backbone instead of an attention backbone in the hypernetwork is detrimental to performance. We speculate that this occurs since a larger hypernetwork incurs overfitting to the training modalities by virtue of being more expressive. Moreover, increasing the context length from 128 to 192 is not beneficial, either, which contradicts our original expectation that larger samples should better approximate the underlying distribution of modality encodings.

\section{Conclusions}

We introduce a novel approach for sample-efficient integration of new modalities (\acronym) into large language models (LLMs). Given a projector, which maps between modality-specific encoders and a decoder LLM, we design a hypernetwork that can adapt it towards any modality. The hypernetwork is trained using data from a limited set of high-resource modalities (e.g., image, audio, and video) and learns to generalise to unseen modalities like satellite images, astronomical images, IMU data, and molecules. We curate a benchmark to measure sample efficiency in this diverse array of modalities, sourcing existing datasets and introducing a new one for galaxy captioning.

We employ isometric transformations to diversify the encoder distributions encountered during training, thereby preventing overfitting. On top of this, grounding modality-specific embeddings on text further enables sample-efficient integration. Overall, \acronym achieves an accuracy comparable with the strongest baseline, namely fine-tuning the shared projector on the new modality, while usually requiring 16$\times$ less labelled data. Finally, we demonstrate the ability of \acronym to generalise to new encoders with arbitrary dimensionality.

By reducing the reliance on large-scale labelled data, our framework facilitates the integration of a diverse set of new modalities into LLMs, expanding the potential applications of multimodal AI models to new areas of geo-location, astronomy, navigation, and biology/medicine. Our approach takes a step forward in the development of truly omni-modal foundation models by extending their coverage to low-resource modalities.

\section{Limitations and Broader Impact} \label{sec:limitations}

Our work assumes that enough modality-specific raw data exists to train an encoder, as it relies on pre-existing, off-the-shelf encoders. This assumption may be violated for some modalities. Our claims are verified only within the precise scope of our setup: generating text only and projecting only one modality at a time. The ethos of this work is facilitating modality integration into LLMs in an efficient and widely accessible way. Our retrofitted models retain the weaknesses of LLMs: similar to text-only settings \cite{zeng-etal-2024-johnny}, modality inputs to MLLMs could be used for jailbreaking \cite{pmlr-v235-gu24e}. Nonetheless, we do not foresee that our work opens more vulnerabilities than existing paradigms.

\subsection*{Acknowledgements}
We gratefully acknowledge Yifu Qiu, Zeyu Huang, Neel Rajani, Andreas Grivas, Giwon Hong, and Alejandro Ariza-Casabona for their invaluable comments and suggestions during the research process. OBİ is supported by the UKRI AI Centre for Doctoral Training (CDT) in Responsible and Trustworthy in-the-world NLP (Grant EP/Y030656/1). AFTM was supported by the project DECOLLAGE (ERC-2022-CoG 101088763) and by FCT/MECI through national funds and when applicable, co-funded EU funds under UID/50008: Instituto de Telecomunicações. EMP is supported by the project AToM-FM (ERC-2025-StG 101222956).

\bibliographystyle{bib}
\bibliography{custom}

\newpage
\appendix
{\LARGE Appendix} 

\section{CAPDELS dataset} \label{sec:capdels}

\begin{figure}[!ht]
  \centering
  \begin{minipage}{0.49\textwidth}
  \includegraphics[width=\linewidth]{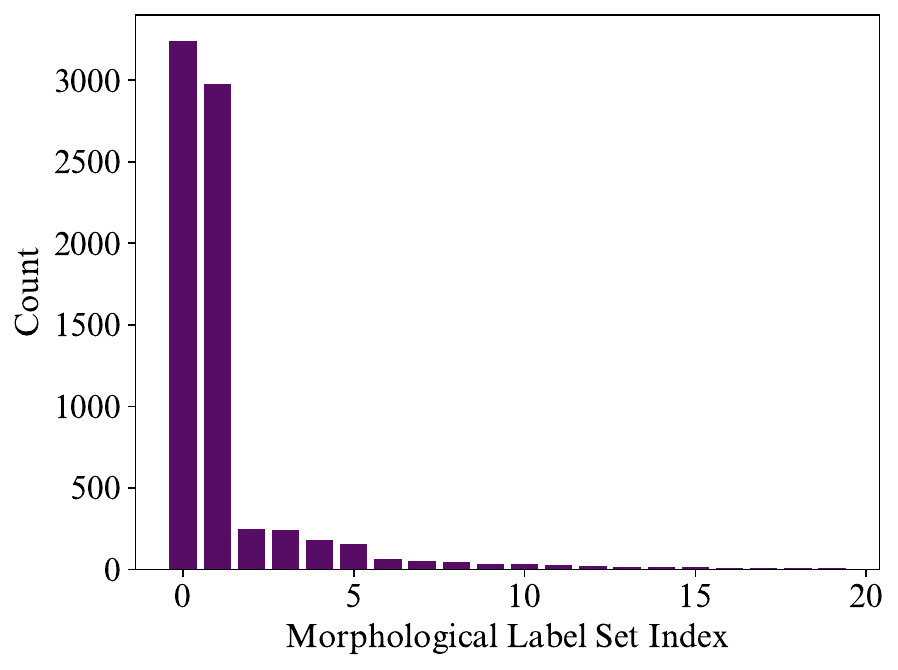}
  \end{minipage}
  \hfill 
  \begin{minipage}{0.49\textwidth}
  \centering
    \includegraphics[width=\linewidth]{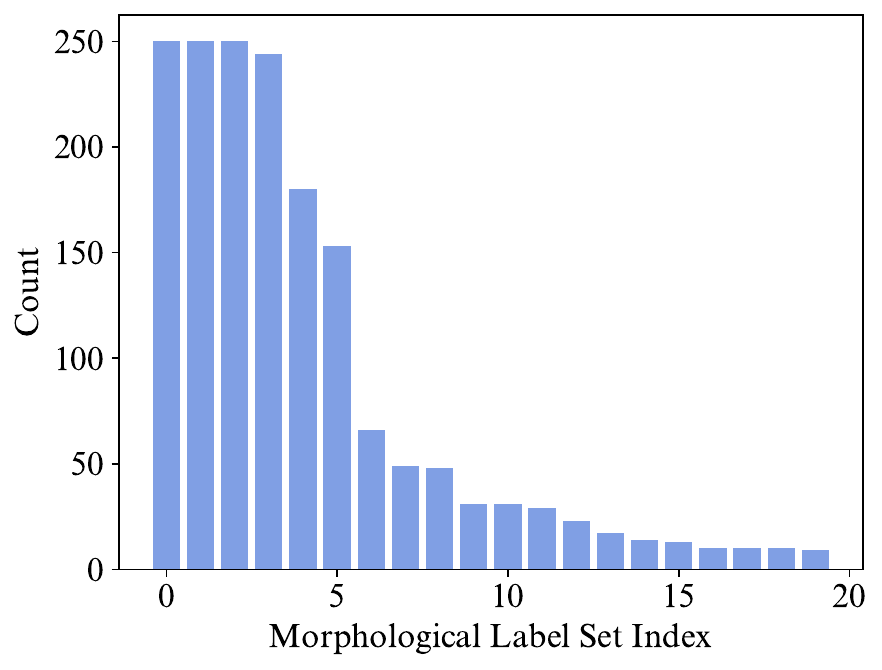}
  \end{minipage}    
  \caption{The frequency of most frequent label sets before (\textbf{left}) and after (\textbf{right}) balancing.}
  \vspace{-0.5em}
  \label{fig:capdels_filtering}
\end{figure}

We introduce CAPDELS, a novel astronomical imaging captioning dataset constructed from the galaxy morphological multi-label classification dataset CANDELS \cite{candels}. While CANDELS contains approximately 50000 examples, only 8000 are `clean' according to author-crafted thresholds. This limited number of clean labels makes it a low-resource dataset, making it a suitable use case for our work. Furthermore, the distribution of morphological label sets is highly imbalanced; only 170 distinct sets exist within these 8000 examples, with approximately 6000 examples belonging to just two dominant sets (see \cref{fig:capdels_filtering}). To address this imbalance, we pruned the most frequent two label sets to match the size of the third-largest set, leaving only 2045 galaxy images in total.

The CANDELS dataset employs a classification tree for categorising galaxy morphology. We leveraged this structure by using the probabilities assigned by annotators for each galaxy to determine its corresponding label sets.  We then used 4-bit quantised Qwen-2.5-32B Instruct \cite{qwen25} to generate captions, providing only the morphological label sets as input – the LLM does not have access to the images themselves. Captions were generated using a system prompt inspired by \cite{liu2023llava}, along with a JSON dictionary containing each sample’s morphological information, producing three captions per image. We used sampling hyperparameters of temperature $= 0.4$, top-k $= 30$, and top-p $= 0.8$ to encourage consistent outputs.

CAPDELS will be released under the same CC BY-NC-SA 4.0 licence as the CANDELS dataset.

\begin{table}[!htbp]
\begin{tcolorbox}[colback=gray!5!white,colframe=black!95!black,title=\small System prompt to Qwen-2.5-32B Instruct for caption generation\label{tab:capdels_sys_prompt}] 
You are an AI assistant tasked with generating a caption for a galaxy image based on its morphological structure. The information about the galaxy will be provided to you as a JSON dictionary, which includes details about its morphological properties. \\

Your job is to create a caption using all the given morphological details from the JSON dictionary. Ensure your caption is: \\

\begin{itemize}
  \item Simple and easy to understand
  \item Concise but specific
  \item Complete (do not omit any details except probabilities)
  \item Using the exact astronomical terminology found in the JSON dictionary
\end{itemize}
\end{tcolorbox}
\end{table}

\begin{table}[!htbp]
\begin{tcolorbox}[colback=gray!5!white,colframe=black!95!black,title=An input example to Qwen-2.5-32B Instruct] 
\begin{Verbatim}[fontsize=\fontsize{8.}{10}\selectfont]
 "Is the galaxy simply smooth and rounded, with no sign of a disk?": 
  "Answer: features or disk. Probabilities: smooth: 15%
   artifact: 0%
 "Does the galaxy have a mostly clumpy appearance?": 
  "Answer: no. Probabilities: yes: 23%
 "Could this be a disk viewed edge-on?": 
  "Answer: no. Probabilities: yes: 7%
 "Is there a sign of a bar feature through the centre of the galaxy?": 
  "Answer: no. Probabilities: yes: 10%
 "Is there any sign of spiral arm pattern?": 
  "Answer: yes. Probabilities: yes: 89%
 "How tightly wound do the spiral arms appear?": 
  "Answer: loose. Probabilities: tight: 35%
 "How many spiral arms are there?": 
  "Answer: 1. Probabilities: 1: 82%
   can't tell: 9%
 "How prominent is the central bulge, compared with the rest of the galaxy?": 
  "Answer: obvious. Probabilities: no bulge: 16%
   dominant: 23%
 "Is the galaxy currently merging or is there any sign of tidal debris?": 
  "Answer: tidal debris. Probabilities: merging: 1%
  both: 2%
}
\end{Verbatim}
\end{tcolorbox}
\end{table}

\begin{table}[!htbp]
\begin{tcolorbox}[colback=gray!5!white,colframe=black!95!black,title=\small Qwen-2.5-32B Instruct generated caption] 
The galaxy has a distinct disk structure with no signs of being edge-on, featuring an obvious central bulge and one loose spiral arm. There is no bar feature through the centre, but there are signs of tidal debris present. The appearance is not clumpy. \\ \rule{\linewidth}{0.4pt} \\ 

$\begin{array}{l}
\includegraphics[width=0.15\linewidth]{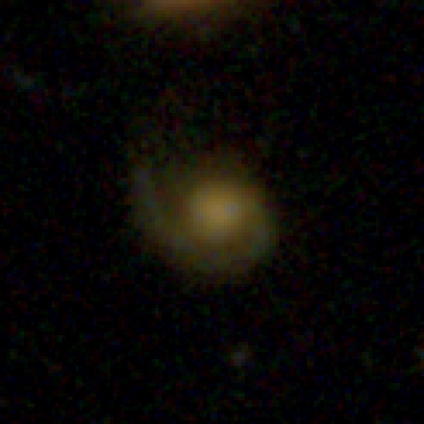}
\end{array}$ The image for reference (not used by LLM).
\end{tcolorbox}
\end{table}

\newpage

\section{Plugging Adapters to All Layers} \label{sec:arch_choice}

\begin{figure}[!ht]
  \centering
  \includegraphics[width=\linewidth]{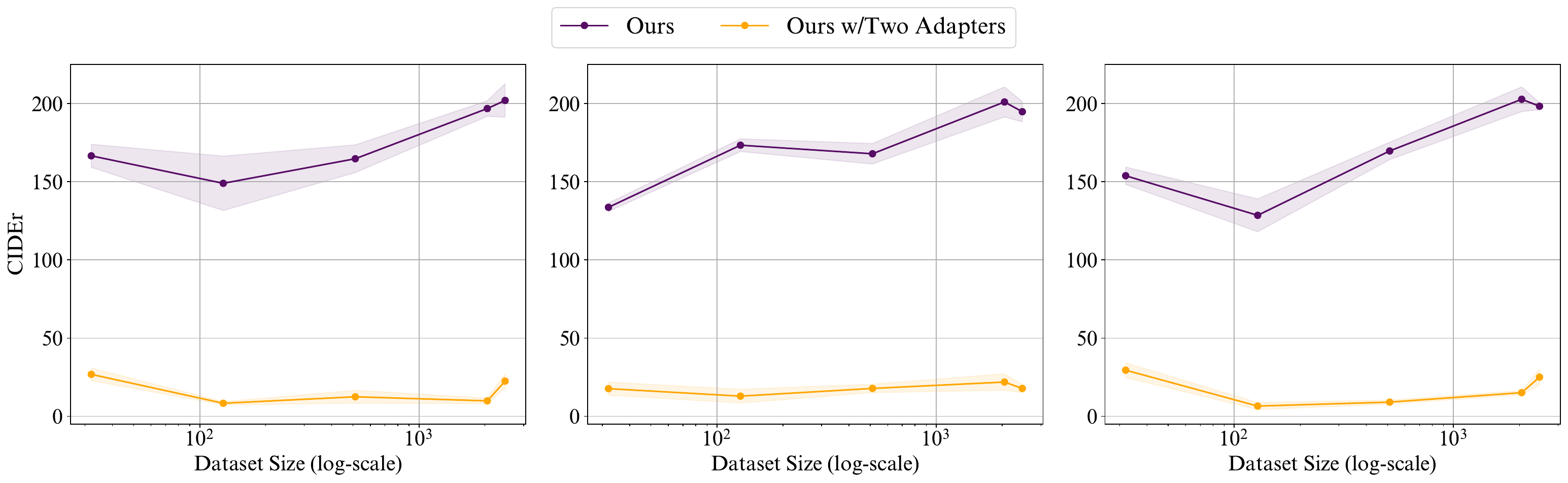}
  \hspace{1em}
  \begin{minipage}[t]{.31\linewidth}
  \vspace{-1em}
  \centering
  \subcaption{ViT-Base-32}\label{fig:sydney_vitb32_2layer}
  \end{minipage}%
  \begin{minipage}[t]{.31\linewidth}
  \vspace{-1em}
  \centering
  \subcaption{ViT-Large-14}\label{fig:sydney_vitl14_2layer}
  \end{minipage}%
  \hspace{1em}
  \begin{minipage}[t]{.31\linewidth}
  \vspace{-1em}
  \centering
  \subcaption{ResNet-50}\label{fig:sydney_rn50_2layer}
  \end{minipage}
  \caption{Comparison of different adapter integration techniques for the SydneyCaptions dataset. 
  }
  \label{fig:sydney_2layer}
\end{figure}

As \cref{fig:sydney_2layer} shows, we encountered a significant performance decrease when training the hypernetwork to add adapters to both projector layers during training. The qualitative results indicate that the projector built using the two-layer setup overfits to the hypernetwork training data (e.g., a drastic increase in usage of training data vocabulary). We hypothesise that bypassing the second layer and adapter during training mitigates this overfitting issue. Nonetheless, we speculate that this effect is due to the relative lack of variation in the hypernetwork and projector training data, and can be overcome by diversifying the training data and increasing its scale.

\section{Ablations for the Hyperparameter Architecture}
\label{app:hyp_abl}

An ablation comparing different design choices for the hypernetwork architecture is shown in \cref{tab:ablations}. Specifically, we consider removing text grounding (\textit{w/o Text}), removing isometric transformations (\textit{w/o IsoTransf}), or both (\textit{w/o Text \& IsoTransf}). We also evaluated a 2-layer Transformer hypernetwork instead of a single attention layer (\textit{Larger Hypernet}) and expanding the hypernetwork context (\textit{w/ Larger Ctx Len}).

{
\setlength{\tabcolsep}{0.4em}
\begin{table}[h!]
\caption{\textbf{Ablations.} The best result for each dataset size is bolded. Multiple methods are bolded if they fall within each other's standard error range. ViT-L-14 and Tiny encoder variants are used for SydneyCaptions and CAPDELS, respectively.}
\label{tab:ablations}
\resizebox{\textwidth}{!}{
\begin{tabular}{@{}lllllll@{\qquad}llllll@{}}
\toprule
 & \multicolumn{6}{c}{\textbf{CAPDELS}}   & \multicolumn{6}{c}{\textbf{SydneyCaptions}}  \\ \midrule
Dataset Size & 32 & 128  & 512  & 2048 & 4344 & Avg.  & 32 & 128  & 512  & 2048 & 2485 & Avg.  \\ \midrule
Ours & {136.8} & \textbf{178.4} & {223.5} & {273.3} & 255.0  & 213.4 & 133.7  & \textbf{173.4} & \textbf{167.9} & \textbf{201.0} & \textbf{194.8}  & 174.1 \\
w/o Text & \textbf{165.4} & \textbf{183.2}  & \textbf{278.0} & \textbf{295.6} & \textbf{303.6} & 245.2 & \textbf{145.2} & 143.8  & 139.0  & 147.0  & 175.9  & 150.2 \\
w/o IsoTransf & 24.7 & 108.2  & 201.4  & {243.6} & {280.3} & 171.6 & \textbf{146.5}  & 158.4  & \textbf{174.4}  & \textbf{201.2} & {181.6} & 172.4 \\
w/o Text \& IsoTransf & 16.2 & 123.2  & 190.1  & 221.8  & {276.1} & 165.5 & \textbf{141.5}  & 158.7  & {157.0} & {167.8} & 155.2  & 156.0 \\ \midrule
w/ Larger Hypernet  & 111.5  & 150.7  & 197.3  & 210.3  & 266.0  & 187.2 & {134.9} & 153.1 & 124.8 & 169.2 & 158.5 & 148.1 \\
w/ Larger Ctx Len &  {143.2} & \textbf{183.7} & {236.0} & 237.5 & 245.2 & 209.1 & 117.5  & 135.7  & 149.7  & 150.8  & 170.9  & 145.0 \\ \bottomrule
\end{tabular}}
\vspace{-1em}
\end{table}}

\section{Comparing Dimensionality Reduction Methods}
\label{app:dim_red}

\begin{figure}[h!]  %
  \centering
    \subfloat{\includegraphics[width=0.65\textwidth]{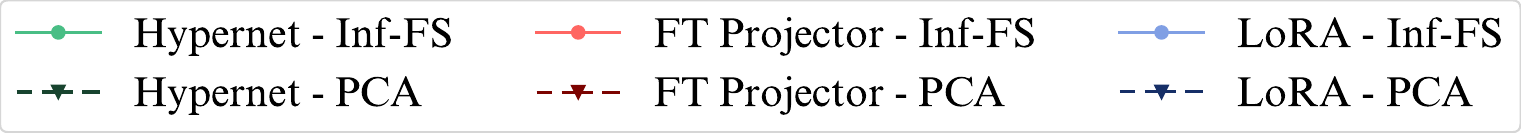}} \\\subfloat{\includegraphics[width=0.48\textwidth]{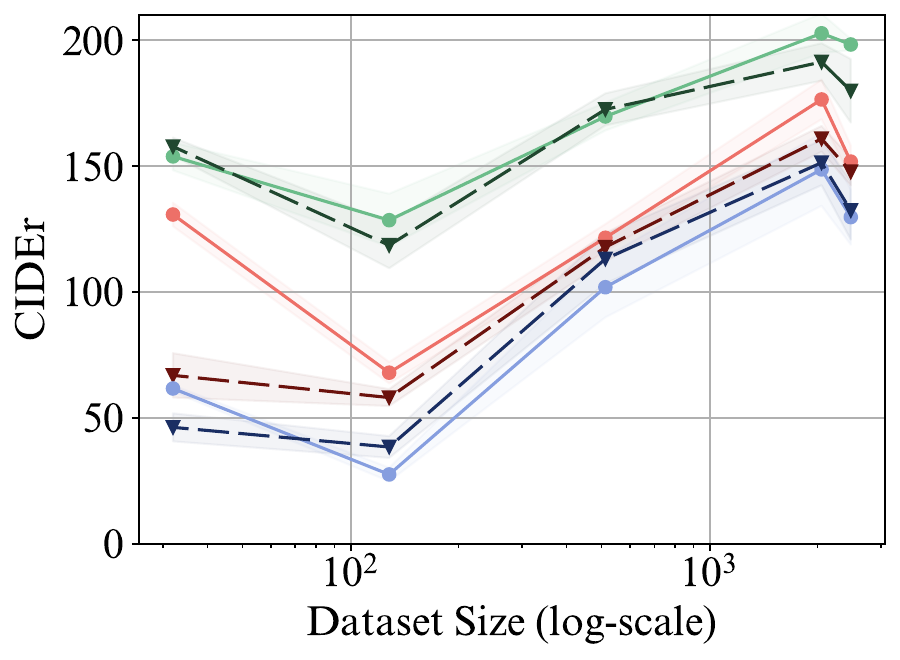}\label{fig:sydney_pca}}
  \hfill\subfloat{\includegraphics[width=0.48\textwidth]{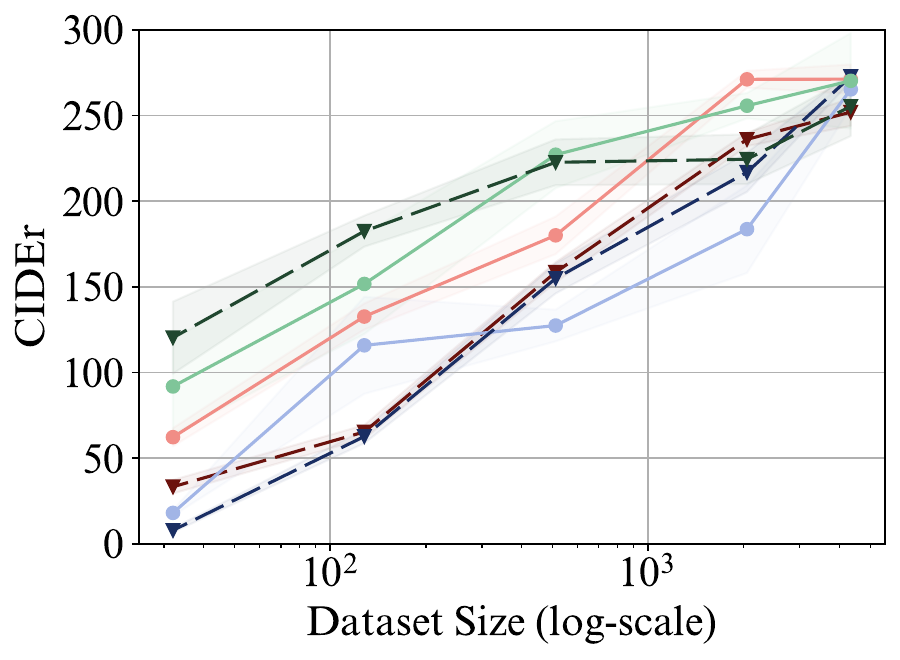}\label{fig:capdels_pca}}
  \caption{Results with PCA and Inf-FS dimensionality reduction techniques for SydneyCaptions (\textbf{left}) and CAPDELS (\textbf{right}) datasets.} %
  \label{fig:all_pca}
\end{figure}

When comparing dimensionality reduction techniques (see \cref{fig:all_pca}), we found that while PCA performed similarly to, and sometimes slightly better than, Inf-FS in hypernetwork and LoRA experiments, its performance significantly degraded, especially on the \textit{FT Projector} baseline. Because PCA offered no consistent advantage over Inf-FS across datasets when used with the hypernetwork, and negatively impacted the performance of \textit{FT Projector}, we report results only for Inf-FS in the main paper.

\section{Comparing Different Adapter Generation Techniques} \label{sec:adapter_generation}

\begin{figure}[htbp]
  \centering
  \begin{minipage}{0.45\textwidth}
  \includegraphics[width=.95\linewidth]{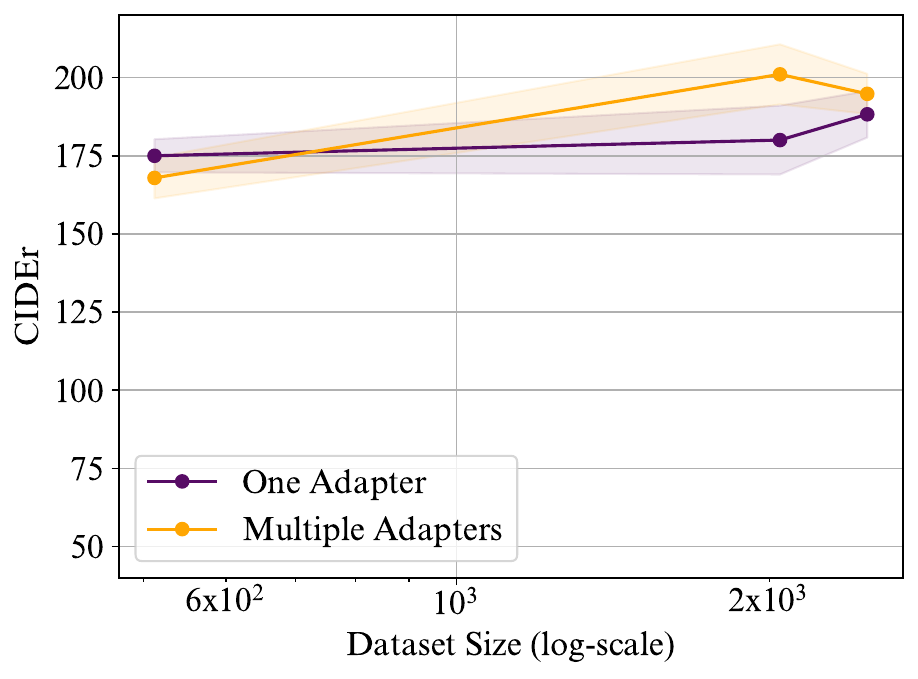}
  \end{minipage}
  \hfill
  \begin{minipage}{0.45\textwidth}
  \includegraphics[width=.95\linewidth]{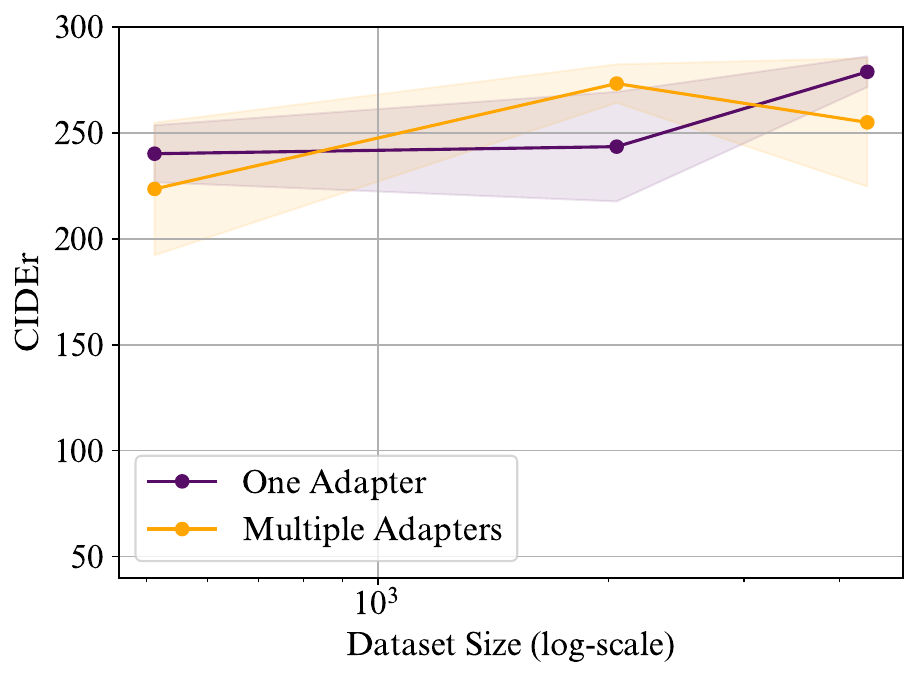}
  \end{minipage}
  \caption{Comparison of single adapter versus averaging multiple adapters for SydneyCaptions - ViT-L-14 \textbf{(left)} and CAPDELS - ConvNeXt-Nano \textbf{(right)} setups. 
  }
  \vspace{-0.5em}
  \label{fig:adapter_techniques}
\end{figure}

In \cref{fig:adapter_techniques}, we compare the single adapter approach with the multiple adapter averaging approach. We observe that generating multiple adapters achieves comparable performance to the single adapter method, while incurring only negligible computational overhead (at most 16 seconds for 206 adapters on the full ChEBI-20 dataset split).

\section{Cross-Modal Similarity of Embeddings}
\label{app:xmod_sim}

\begin{figure}[!ht]
  \centering
  \includegraphics[width=0.59\linewidth]{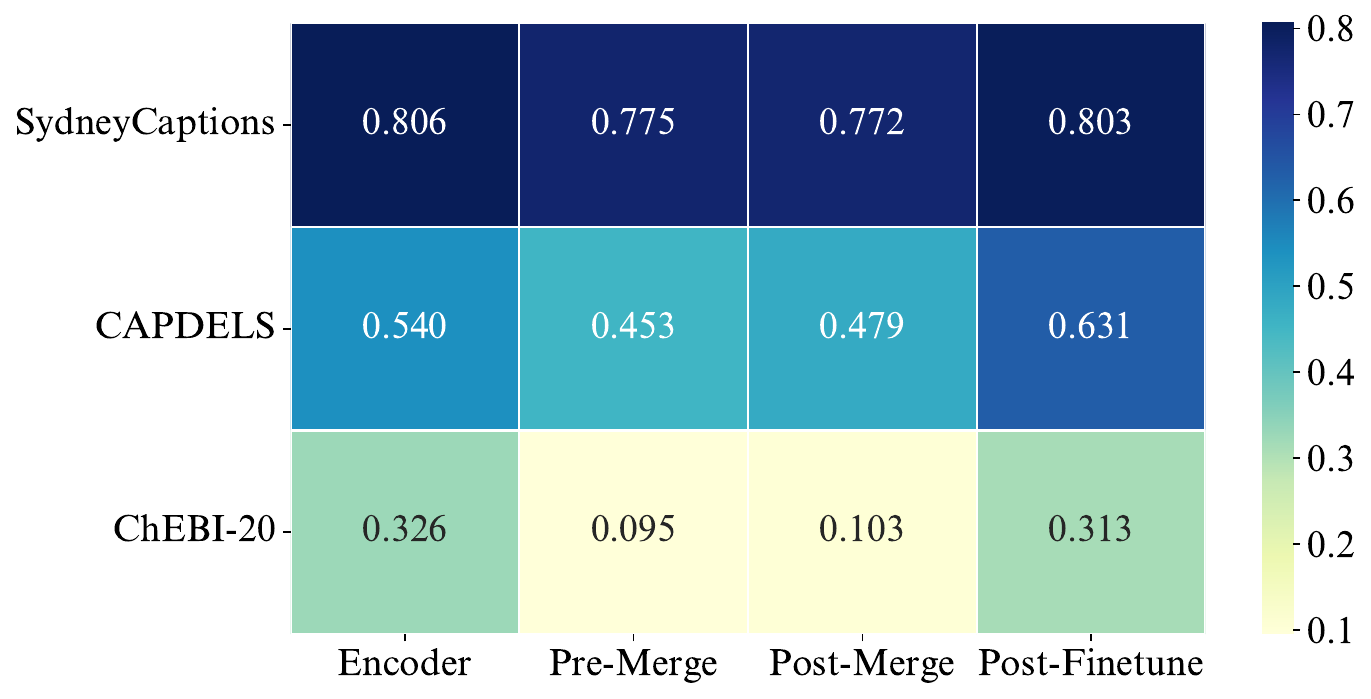}
  \caption{Linear CKA scores between modality embeddings at different stages and text embeddings. `Encoder' embeddings are unprocessed. Embeddings projected with the pre-trained projector before and after merging with the adapters are labelled `Pre-Merge' and `Post-Merge', respectively. `Post-Finetune' embeddings are extracted after fine-tuning of the merged projector. 2048 samples are used for ChEBI-20 and 128 for others.}
  \label{fig:chebi20_n_cka}
\vspace{-0.5em}
\end{figure}

We analyse the effect of different stages of our pipeline on the similarity between the embeddings of each unseen modality and the corresponding text. The similarity scores are obtained with Linear CKA \cite{cka} and shown in \cref{fig:chebi20_n_cka} in the form of a heatmap.
We find that the embeddings of text and each unseen modality are originally well aligned, but the modality embeddings are distorted during mapping into the LLM input space. However, the amount of distortion is reduced progressively as we incorporate adapters and further fine-tune the projector. Moreover, the effect of these two steps is more pronounced in the galaxy modality, as galaxy encoders lack text conditioning. The satellite encoders are already well aligned, which might explain the comparatively higher performance of training a projector from scratch. The molecule modality is arguably the most unique and challenging modality, which might explain the smaller effect of adapter merging on embedding alignment.

\section{Qualitative Examples} \label{sec:qual_examples}

\begin{table}[!htbp]
    \centering
    \small
      \caption{Qualitative examples for SydneyCaptions dataset and ViT-L-14 encoder for methods trained with 128 samples}
    \begin{tabular}{ @{}p{0.18\linewidth}p{0.24\linewidth}p{0.52\linewidth}@{}}
     \toprule
      \multicolumn{1}{c}{\textbf{Image}} & \multicolumn{1}{c}{\textbf{Ground Truth}} & \multicolumn{1}{c}{\textbf{Predictions}} \\ 
    \cmidrule(r){1-1}\cmidrule(lr){2-2}\cmidrule(l){3-3}
     \raisebox{-\totalheight}{\includegraphics[width=\linewidth]{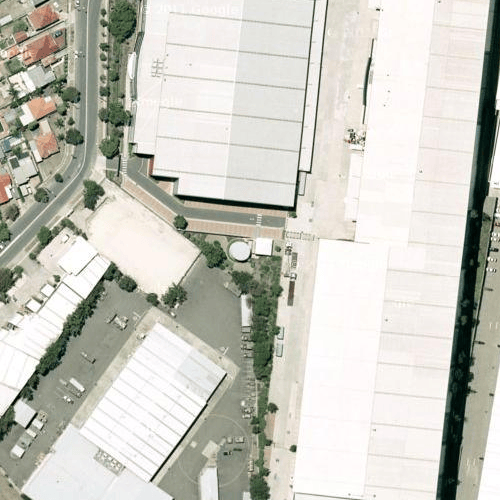}}
      & \begin{itemize}[topsep=0pt, leftmargin=*]
          \item[] This is an industrial area with many white buildings densely arranged while a residential area beside
      \end{itemize}
      & 
      \begin{itemize}[topsep=0pt, leftmargin=*]
        \setlength\itemsep{0.3em}
      \item[] \textbf{Ours:} This is an industrial area with some roads and many buildings
      \item[] \textbf{FT Proj:} This image shows a residential area with some buildings and some roads
      \item[] \textbf{Proj:} Some buildings there
      \item[] \textbf{LoRA:} There is a residential area with some houses on a go road
      \end{itemize}
      \\ 
\raisebox{-\totalheight}{\includegraphics[width=\linewidth]{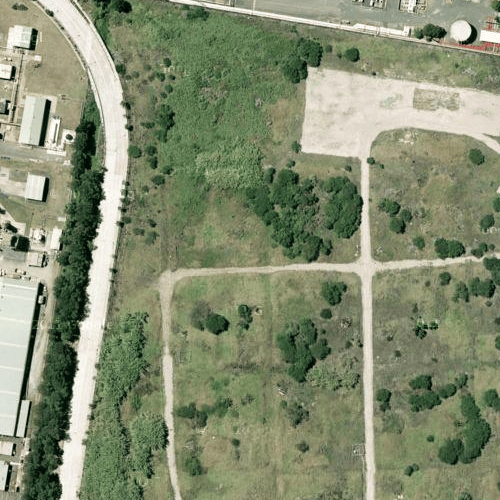}}
      & \begin{itemize}[topsep=0pt, leftmargin=*]
          \item[] This is a meadow with some green bushes on it while some roads passed by
      \end{itemize}
      & 
      \begin{itemize}[topsep=0pt, leftmargin=*]
        \setlength\itemsep{0.3em}
      \item[] \textbf{Ours:} This is a big meadow with some roads on it 
      \item[] \textbf{FT Proj:} There are some white flowers and some white sidewalks on the green field
      \item[] \textbf{Proj:} Some green bushes and white sand on the beach
      \item[] \textbf{LoRA:} Some sections of a farm are covered with green grass while others are divided by a straight road
      \end{itemize}
      \\ 
      \bottomrule
      \end{tabular}
      \label{tab:sydney_qual}
\end{table}

In Table \ref{tab:sydney_qual}, we observe that the hypernetwork demonstrates a stronger ability to ground tasks compared to the fine-tuned projector, consistently generating correct outputs even if some details are omitted. While the fine-tuned projector often produces answers close to the ground truth, it occasionally makes mistakes and visibly retains remnants of its pre-trained state in the generated words---suggesting it has not fully adapted to the new task. \textit{LoRA} frequently follows the \textit{FT Projector}, omitting the industrial aspect of the image as well as hallucinating a ``go road''. This improved task grounding makes it applicable across diverse modalities, rather than simply memorising pre-training data. 

Qualitative results demonstrate that the hypernetwork excels at task adaptation compared to other methods. Specifically, the hypernetwork consistently generates correct answers verbatim, while alternatives often introduce word choices absent from the training data, a clear sign of insufficient detachment from pre-trained knowledge. Furthermore, all of the competing approaches except \textit{LoRA} failed to identify the merger occurring in the first image (as shown in \cref{tab:capdels_qual}), highlighting a limitation in their ability to perceive key details. Even when methods like \textit{FT Projector} produce seemingly reasonable outputs, they tend to go beyond the specified task description, and the \textit{Projector} and \textit{LoRA} generations, while reflecting some truth, frequently hallucinate by adding irrelevant information such as incorrect details about ``comfort'', ``presence'', or colour.

In \cref{tab:sensorcaps_qual}, we observe that the hypernetwork correctly answers the questions with a similar format to the ground truth while other baselines struggle to do so. In the first example, \textit{FT Projector} incorrectly identifies the rotation as minimal, while other baselines produce nonsensical and unrelated outputs. In the second example, our approach is completely correct while the second-best \textit{LoRA} baseline incorrectly hallucinates that the accelerometer data ranges from ``-1 to 1 g'' even though it stays the same -- failing to finish the description within the maximum token limit.

While hypernetworks offer the most promising approach for integrating new modalities into LLMs, their generated responses are not without flaws; we observed errors such as misidentifying proanthocyanidins as cinnamaldehydes and incorrectly classifying GABA as a glycerophospholipid (see \cref{tab:chebi20_qual}). The \textit{FT Projector} baseline performs similarly -- correctly identifying proanthocyanidin as a polyphenol, for example -- but struggles with accurately listing the corresponding tree species. \textit{LoRA} baseline captures salient patterns of the molecule, although it eventually provides non-factual information. Despite these limitations, the \textit{Projector} baseline frequently exhibits significant inaccuracies. Importantly, our findings suggest that integrating more complex, knowledge-intensive modalities requires larger training datasets; however, we demonstrate that leveraging information from other modalities can significantly ease this integration process.

\begin{table}[!htbp]
    \centering
    \small
      \caption{Qualitative examples for CAPDELS dataset and ConvNeXt-Tiny encoder for methods trained with 128 samples}
    \begin{tabular}{ @{}p{0.18\linewidth}p{0.22\linewidth}p{0.53\linewidth}@{}}
     \toprule
      \multicolumn{1}{c}{\textbf{Image}} & \multicolumn{1}{c}{\textbf{Ground Truth}} & \multicolumn{1}{c}{\textbf{Predictions}} \\ 
    \cmidrule(r){1-1}\cmidrule(lr){2-2}\cmidrule(l){3-3}
     \raisebox{-\totalheight}{\includegraphics[width=\linewidth]{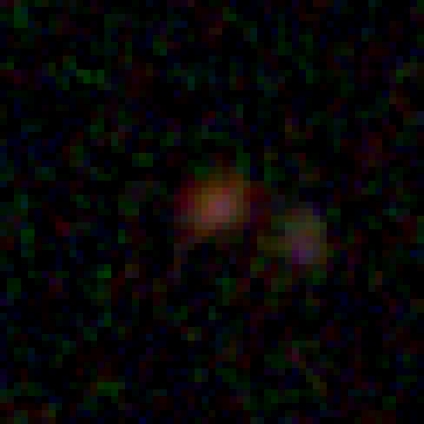}}
      & \begin{itemize}[topsep=0pt, leftmargin=*]
          \item[] The galaxy is smooth and rounded, with a shape that is in between completely round and cigar-shaped. There are signs that the galaxy is currently merging with another galaxy.
      \end{itemize}
      & 
      \begin{itemize}[topsep=0pt, leftmargin=*]
        \setlength\itemsep{0.3em}
      \item[] \textbf{Ours:} The galaxy is smooth and rounded, with a shape that is in between completely round and cigar-shaped. There are signs that the galaxy is currently merging with another galaxy.
      \item[] \textbf{FT Proj:} The galaxy is smooth and rounded with no signs of a disk, having an immediate and sharp appearance.
      \item[] \textbf{Proj:} The galaxy is smooth and rounded, with no signs of a disk. The galaxy is a small, showing no signs of merging or being in a state of merger. There are no signs that it has undergone any histological changes.
      \item[] \textbf{LoRA:} The galaxy is smooth and rounded with no signs of a disk, having an in-between roundness, and shows signs of being present. There are indications that the galaxy is currently undergoing the process.
      \end{itemize}
      \\ 
\raisebox{-\totalheight}{\includegraphics[width=\linewidth]{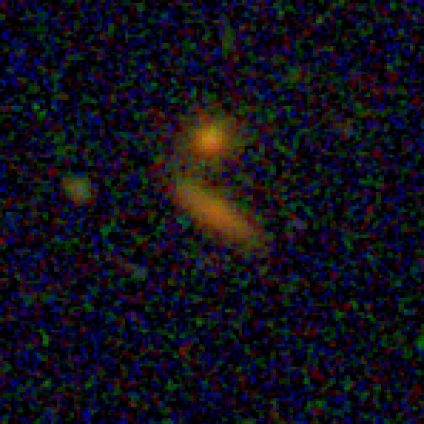}}
      & \begin{itemize}[topsep=0pt, leftmargin=*]
          \item[] The galaxy is smooth and rounded, with a cigar-shaped appearance, and shows no signs of merging or tidal debris.
      \end{itemize}
      & 
      \begin{itemize}[topsep=0pt, leftmargin=*]
        \setlength\itemsep{0.3em}
      \item[] \textbf{Ours:} The galaxy is smooth and rounded, with a cigar-shaped appearance and shows no signs of merging or tidal debris.
      \item[] \textbf{FT Proj:} The galaxy is smooth and evenly distributed, showing no signs of merging or tidal debris. The galaxy appears to be completely unrelated to others in the image.
      \item[] \textbf{Proj:} The galaxy appears smooth and rounded, with a cigar-shaped appearance and has a peach-colored hue.
      \item[] \textbf{LoRA:} The galaxy is smooth and rounded, with a cigar-shaped appearance, and shows no signs of merging or tidal debris. It appears more comfortable and has no indication of merging or tidal debris.
      \end{itemize}
      \\ 
      \bottomrule
      \end{tabular}
      \label{tab:capdels_qual}
\end{table}

\begin{table}[!ht]
    \centering
    \small
      \caption{Qualitative examples for SensorCaps dataset and LIMU-BERT for methods trained with 128 samples}
    \begin{tabular}{ @{}p{0.34\linewidth}p{0.63\linewidth}@{}}
     \toprule
      \multicolumn{1}{c}{\textbf{Ground Truth}} & \multicolumn{1}{c}{\textbf{Predictions}} \\ 
    \cmidrule(r){1-1}\cmidrule(l){2-2}
    Features: Walking, moderate intensity Narration: 1. The accelerometer data shows varying values along the x, y, and z axes, suggesting movement in multiple directions. 2. In the gyroscope data, there are consistent fluctuating values on all axes, indicating rotational movements from walking. 3. The positive and negative values in both sensor data indicate changes in acceleration and angular velocity during the walking activity. 4. The peaks and troughs in the data suggest steps taken in a rhythmic pattern.
      & 
      \begin{itemize}[topsep=0pt, leftmargin=*]
        \setlength\itemsep{0.3em}
      \item[] \textbf{Ours:} In the gyroscope data, there are varying values of x, y, and z axes. The gyroscope data also shows fluctuating values of z, x, and y axes. The user is likely rotating their body while the gyroscope and accelerometer data capture the changes in orientation and movement. The accelerometer data may indicate changes in body position and movement. The fluctuations in gyroscope data suggest the user is moving or rotating in different directions.
      \item[] \textbf{FT Proj:} Features in the description of the gyroscope readings suggest that the device is moving in a consistent pattern. The magnitude of the gyroscope data is not significant, indicating minimal rotation. The axis values are relatively close in value, indicating a stable movement in all axes. The accelerometer data shows minimal movement in all axes, consistent with a stationary device.
      \item[] \textbf{Proj:} None, None, None, None.
      \item[] \textbf{LoRA:} However, the accuracy is low. 
      \end{itemize} \\ 
       Features: Stationary, subtle movements. Narration: The gyroscope data shows minimal variations around 0, indicating little rotational movement. Accelerometer data remains around 9.8 m/s², denoting no significant changes in acceleration. The consistent values suggest the user is standing still. Slight fluctuations in gyroscope and accelerometer readings imply subtle swaying or adjustments in posture while standing.
      & 
      \begin{itemize}[topsep=0pt, leftmargin=*]
        \setlength\itemsep{0.3em}
      \item[] \textbf{Ours:} The gyroscope data shows minimal fluctuations, indicating minimal movement. The accelerometer data indicates consistent values, suggesting the user is standing still. The user's posture is likely stationary, as the gyroscope readings are close to zero and the accelerometer values are near the average weight. The slight variations in the accelerometer data may be attributed to minor adjustments in posture or slight movements while standing.
      \item[] \textbf{FT Proj:} Note: No additional context is provided.
      \item[] \textbf{Proj:} None, it's a non-physical activity.
      \item[] \textbf{LoRA:} The symptoms match the characteristics of a person standing still with minimal physical activity. Here's a description of the gyroscope and accelerometer data:

    1. Gyroscope data:
    - The x, y, and z axes show minimal changes, indicating no significant rotation or movement in these planes. The data is relatively stable, with slight variations, suggesting a stationary position.
    
    2. Accelerometer data:
    - The x, y, and z axes show consistent changes, typically ranging from -1 to 1 g. This
    
      \end{itemize} \\ 
      \bottomrule
      \end{tabular}\vspace{-1em}
      \label{tab:sensorcaps_qual}
\end{table}

\begin{table}[!ht]
    \centering
    \small
      \caption{Qualitative examples for ChEBI-20 dataset and MolCA for methods trained with 2048 samples}
    \begin{tabular}{ @{}p{0.27\linewidth}p{0.70\linewidth}@{}}
     \toprule
      \multicolumn{1}{c}{\textbf{Ground Truth}} & \multicolumn{1}{c}{\textbf{Predictions}} \\ 
    \cmidrule(r){1-1}\cmidrule(l){2-2}
      \begin{itemize}[topsep=0pt, leftmargin=*]
      \item[] \textbf{2D structure image:}
          \item[] \raisebox{-\totalheight}{\includegraphics[width=\linewidth]{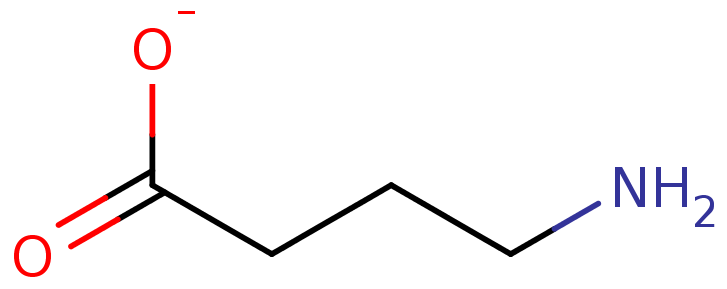}}
          \item[] \textbf{Description:} The molecule is a gamma-amino acid anion resulting from the deprotonation of the carboxy group of gamma-aminobutyric acid. It has a role as a human metabolite. It derives from a butyrate. It is a conjugate base of a gamma-aminobutyric acid.
      \end{itemize}
      & 
      \begin{itemize}[topsep=0pt, leftmargin=*]
        \setlength\itemsep{0.3em}
      \item[] \textbf{Ours:} She is a species to which a molecule is a gamma-aminobutyric acid (GABA) conjugate that has been metabolized to a glycerol-3-phosphate conjugate. It is a glycerophospholipid and a member of the gamma-aminobutyric acid (GABA) conjugate family. It has a role as a human metabolite and a human metabolite. It is a major player in the metabolism of GABA. It is a conjugate acid of a gamma-aminobutyric acid (GABA) molecule.
      \item[] \textbf{FT Proj:} The molecule is a tetracyclic compound that consists of a 2,3,5-trisubstituted cyclohexene with a 2,3,5-trisubstituted 2,3,5-trimethylcyclohexene-1,4,5-diol and 2,3,5-trisubstituted 2,3,5-trimethylcyclohexene-1,4,5-dione. It has a role as a plant metabolite and a polyphenol. It derives from a 2,3,5-trisubstituted cyclohexene and a 2,3,5-trisubstituted 2,3,5-trimethylcyclohexene-1,4,5-diol.
      \item[] \textbf{Proj:} The molecule is a 2,3-diphenyl-2-pyridine-5-oxide derivative that is similar to 2,3-diphenyl-2-pyridine-5-oxide.
      \item[] \textbf{LoRA:} The molecule is a glucuronic acid derivative; it is a conjugate acid of 3-deoxy-D-glucuronic acid. It has a role as a human metabolite and a metabolite. It derives from a 3-deoxy-D-glucuronic acid. It is a conjugate acid of a 3-deoxy-D-glucuronic acid.
      \end{itemize} \\ 
\begin{itemize}[topsep=0pt, leftmargin=*]        
      \item[] \textbf{2D structure image:}
          \item[] \raisebox{-\totalheight}{\includegraphics[width=\linewidth]{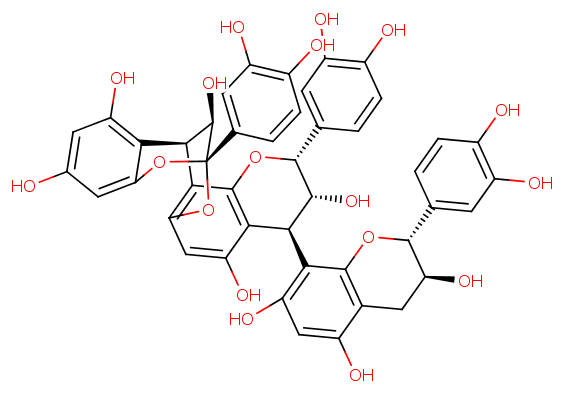}}
          \item[] \textbf{Description:} The molecule is a proanthocyanidin found in Cinnamomum cassia. It has a role as a cyclooxygenase 2 inhibitor and a plant metabolite.
      \end{itemize}
      & 
      \begin{itemize}[topsep=0pt, leftmargin=*]
        \setlength\itemsep{0.3em}
      \item[] \textbf{Ours:} The molecule is a type of cinnamaldehyde found in the leaves of Cinnamomum aromaticum (Cinnamomum zeylanicum). It has a role as a plant metabolite and a cinnamaldehyde.
      \item[] \textbf{FT Proj:} The molecule is a polyphenol that is a secondary metabolite of the tree species Pinus spp. It is found in the bark of Pinus spp. and is composed of a mixture of various flavonoids and phenolic compounds.
      \item[] \textbf{Proj:} The molecule is a prostanoid derived from 6-deoxy-5,6,7,8-tetrahydro-2-methoxy-5,6,7,8-(2-methyl-9,10-phenyl)-2H-cyclopropeno-1,4,7,8(3,5)-tetraones. The molecule is a prostanoid and a flavonoid present in the plant Pterocarpus spinosus. It has a role as a flavonoid and a plant extract used in traditional medicine.
      \item[] \textbf{LoRA:} Alicin is a metabolite of aloe vera, a plant-based compound that is used in the treatment of hemorrhoids and other hemorrhagic conditions. It has a role as a metabolite and an ethyl ester.
      \end{itemize} \\ 
      \bottomrule
      \end{tabular}\vspace{-1em}
      \label{tab:chebi20_qual}
\end{table}

\clearpage

\section{Pseudocode} \label{sec:pseudocodes}

\begin{algorithm}
\renewcommand{\algorithmicrequire}{\textbf{Input:}}
\renewcommand{\algorithmicensure}{\textbf{Output:}}
\makeatletter
\xpatchcmd{\algorithmic}{\itemsep\z@}{\itemsep=1.5pt}{}{}
\makeatother
\caption{Projector pre-training}
\label{alg:proj_pretrain}
\begin{algorithmic}[1] %

\Require Modality encoders \(\{\text{enc}_m\}_{m=1}^M\), LLM, training datasets \( \{\mathcal{D}_{m}\}_{m=1}^M\), dataset instructions \(\{\mathbf{i}_{m}\}_{m=1}^M\), projector \(\text{proj}\) with parameters \(\psi\), initial projector parameters \(\psi_{\text{init}}\), cross-entropy $H(\cdot)$   

\Ensure Pre-trained projector parameters \(\psi^\star\)

\State \(\psi \leftarrow \psi_\text{init}\)

\While{not converged}
    \State \(m \sim \{1, \dots, M\}\) \Comment Randomly sample a modality
    
    \State \(\mathbf{x}, \mathbf{y} \sim \mathcal{D}_{m}\)\Comment Get modality input and text output
        
    \State $\mathbf{z} \leftarrow \text{proj}_\psi(\text{enc}_m({\mathbf{x}})) \oplus \mathbf{i}_m$

    \State \(\ell \leftarrow H(\text{LLM}(\mathbf{z}), \mathbf{y})\) \Comment Calculate cross-entropy loss

    \State Update \(\psi\) using \(\nabla_{\psi}\) w.r.t. \(\ell\) 
\EndWhile
\State \(\psi^\star \leftarrow \psi\)
\end{algorithmic}
\end{algorithm}

\begin{algorithm}
\renewcommand{\algorithmicrequire}{\textbf{Input:}}
\renewcommand{\algorithmicensure}{\textbf{Output:}}
\makeatletter
\xpatchcmd{\algorithmic}{\itemsep\z@}{\itemsep=2pt}{}{}
\makeatother
\caption{Hypernetwork training}
\label{alg:hypernet_training_simple}
\begin{algorithmic}[1] %

\Require Modality encoders \(\{\text{enc}_m\}_{m=1}^M\), text encoder \(\text{enc}_{\text{text}}\), LLM, training datasets \(\{\mathcal{D}_m\}_{m=1}^M\), hypernet $\text{hyp}$ with parameters \(\theta\), pre-trained projector \(\text{proj}\) with parameters \(\psi^\star\), dataset instruction sets \(\{\mathcal{I}_{m}\}_{m=1}^M\), initial hypernet parameters \(\theta_{\text{init}}\), cross-entropy $H(\cdot)$ 

\Ensure Trained hypernetwork parameters \(\theta^\star\)

\State \(\theta \leftarrow \theta_\text{init}\)

\While{not converged}
    \State \(m \sim \{1, \dots, M\}\) \Comment Randomly sample a modality

    \State \(\mathbf{i}_m \sim \mathcal{I}_{m}\)

    \State \(\{\mathbf{x}_{\text{hyp}}, \mathbf{y}_{\text{hyp}}\}_{1}^{S} \sim \mathcal{D}_{m}\)
    
    \State \(\mathbf{x}_{\text{LLM}}, \mathbf{y}_{\text{LLM}} \sim \mathcal{D}_{m}\)
        \State $\bm{Q} \sim \operatorname{Haar}(\mathrm{O}(h_{\text{enc}_m}))$ \Comment{Sample orthogonal matrix}

    \State \(\mathbf{m}_{\text{LLM}} \leftarrow \bm{Q} \, \text{enc}_m(\mathbf{x}_{\text{LLM}})\)
    \State \(\mathbf{m}_{\text{hyp}} \leftarrow \bm{Q} \,\text{enc}_m(\mathbf{x}_{\text{hyp}})\) \Comment{Extract embeddings with orthogonal transform}
    \State \(\delta \leftarrow \text{hyp}_\theta\left(\text{enc}_\text{text}(\mathbf{i}_m) \oplus [\mathbf{m}_{\text{hyp}} \oplus \text{enc}_\text{text}(\mathbf{y}_{\text{hyp}})]_1^S\right)\)
\Comment Generate LoRA with hypernetwork

    \State \(\psi^\prime\leftarrow \psi^\star + \frac{\alpha}{r} \delta \) \Comment Combine projector and LoRA parameters

    \State \(\mathbf{z} \leftarrow \text{proj}_{\psi^\prime}(\mathbf{m}_{\text{LLM}})) \oplus \mathbf{i}_m\)

    \State \(\ell \leftarrow H(\text{LLM}(\mathbf{z}),\mathbf{y}_{\text{LLM}})\) \Comment Calculate loss

    \State Update \(\theta\) using \(\nabla_{\theta}\) w.r.t. \(\ell\) 
\EndWhile

\State \(\theta^\star \leftarrow \theta\)

\end{algorithmic}
\end{algorithm}

\begin{algorithm}
\renewcommand{\algorithmicrequire}{\textbf{Input:}}
\renewcommand{\algorithmicensure}{\textbf{Output:}}
\makeatletter
\xpatchcmd{\algorithmic}{\itemsep\z@}{\itemsep=1.5pt}{}{}
\makeatother
\caption{Few-shot adaptation}
\label{alg:hypernet_adaptation_simple}
\begin{algorithmic}[1] %

\Require Test modality encoder \(\text{enc}_m\), text encoder \(\text{enc}_{\text{text}}\), LLM, training dataset \(\mathcal{D}_m\), trained hypernet hyp with parameters \(\theta^\star\), pre-trained projector proj with parameters \(\psi^\star\), instruction \(\mathbf{i}_m\), cross-entropy $H(\cdot)$   

\Ensure Projector parameters \(\psi^\prime\)

\State $\Delta \leftarrow \varnothing$ 

\For{\(\{\mathbf{x}_{\text{hyp}}, \mathbf{y}_{\text{hyp}}\}_{1}^{S} \in \mathcal{D}_{m}\)}
    \State \(\delta \leftarrow \text{hyp}_{\theta^\star} \left( \text{enc}_\text{text}(\mathbf{i}_m) \oplus [\text{enc}_{m}(\textbf{x}_{\text{hyp}}) \oplus \text{enc}_\text{text}(\mathbf{y}_{\text{hyp}})]_1^S\right)\)
\Comment Generate LoRA

    \State $\Delta \leftarrow \Delta \cup \delta$
\EndFor

\State \(\psi^\prime \leftarrow \psi^\star + \frac{\alpha}{r}\bar{\Delta}\) \Comment Combine projector parameters and averaged LoRAs

\While{not converged}
    \State \(\mathbf{x}_{\text{LLM}}, \mathbf{y}_{\text{LLM}} \sim \mathcal{D}_{m}\)

    \State \(\mathbf{z} \leftarrow \text{proj}_{\psi^\prime}(\text{enc}_m({\mathbf{x}_\text{LLM}})) \oplus \mathbf{i}_m\)

    \State \(\ell \leftarrow H(\text{LLM}(\mathbf{z}),\mathbf{y}_{\text{LLM}})\) \Comment Calculate loss

    \State Update \(\psi^\prime\) using \(\nabla_{\psi^\prime}\) w.r.t. \(\ell\) 
\EndWhile

\end{algorithmic}
\end{algorithm}

\newpage

\section{Results with All Metrics} \label{sec:add_metrics}

We used the \texttt{allenai/scibert\_scivocab\_uncased} \cite{scibert, wolf-etal-2020-transformers} tokeniser for the ChEBI-20 dataset following \cite{molt5}, whereas we used a whitespace tokeniser (or metric-specific tokenisers) for calculating the metrics of the remaining datasets.

\subsection{Llama 3.1 8B Instruct Results}
\label{sec:add_metrics_8b}
\subsubsection{SydneyCaptions dataset} \label{sec:add_mtrc_syd_8b}

\begin{table}[!htbp]
    \centering
    \caption{All results and metrics for the ViT-B-32 encoder on the SydneyCaptions dataset. We show the mean ± standard error calculated over three random seeds.}
\resizebox{\textwidth}{!}{
\begin{tabular}{llrrrrrr}
\toprule
\textbf{Sample Size} & \textbf{Setup} \textbackslash{} \textbf{Metric} & \multicolumn{1}{c}{\textbf{ROUGE-1}} & \multicolumn{1}{c}{\textbf{ROUGE-2}} & \multicolumn{1}{c}{\textbf{ROUGE-L}} & \multicolumn{1}{c}{\textbf{BLEU}} & \multicolumn{1}{c}{\textbf{METEOR}} & \multicolumn{1}{c}{\textbf{CIDEr}} \\
\midrule
\multirow[c]{4}{*}{\hspace{0.5em}32} & Ours & 59.23 $\pm$ 0.41 & 44.73 $\pm$ 0.81 & 56.84 $\pm$ 0.55 & 42.99 $\pm$ 1.29 & 56.42 $\pm$ 0.71 & 163.93 $\pm$ 2.66 \\
 & FT Projector & 54.76 $\pm$ 1.83 & 37.82 $\pm$ 2.32 & 51.92 $\pm$ 2.07 & 36.24 $\pm$ 2.10 & 50.01 $\pm$ 2.01 & 146.25 $\pm$ 8.50 \\
 & Projector & 58.25 $\pm$ 1.24 & 42.11 $\pm$ 1.28 & 55.39 $\pm$ 1.43 & 40.24 $\pm$ 2.91 & 52.72 $\pm$ 0.65 & 155.26 $\pm$ 9.86 \\
 & LoRA & 45.68 $\pm$ 0.27 & 26.39 $\pm$ 1.22 & 42.43 $\pm$ 0.67 & 23.85 $\pm$ 1.98 & 40.51 $\pm$ 1.10 & 79.76 $\pm$ 9.10\vspace{0.75em}\\

\multirow[c]{4}{*}{\hspace{0.5em}128} & Ours & 63.67 $\pm$ 0.39 & 47.48 $\pm$ 0.67 & 60.95 $\pm$ 0.51 & 42.95 $\pm$ 0.41 & 61.97 $\pm$ 0.67 & 184.61 $\pm$ 4.76 \\
 & FT Projector & 63.12 $\pm$ 1.13 & 46.68 $\pm$ 1.36 & 60.28 $\pm$ 1.15 & 43.43 $\pm$ 1.69 & 60.29 $\pm$ 1.09 & 176.34 $\pm$ 6.99 \\
 & Projector & 60.91 $\pm$ 1.04 & 44.62 $\pm$ 0.96 & 57.85 $\pm$ 1.13 & 39.68 $\pm$ 1.79 & 58.90 $\pm$ 0.87 & 155.26 $\pm$ 5.73 \\
\bottomrule
\end{tabular}}
\end{table}

\begin{table}[!htbp]
    \centering
    \caption{All results and metrics for the ViT-L-14 encoder on the SydneyCaptions dataset. We show the mean ± standard error calculated over three random seeds.}
\resizebox{\textwidth}{!}{\begin{tabular}{llrrrrrr}
\toprule
\textbf{Sample Size} & \textbf{Setup} \textbackslash{} \textbf{Metric} & \multicolumn{1}{c}{\textbf{ROUGE-1}} & \multicolumn{1}{c}{\textbf{ROUGE-2}} & \multicolumn{1}{c}{\textbf{ROUGE-L}} & \multicolumn{1}{c}{\textbf{BLEU}} & \multicolumn{1}{c}{\textbf{METEOR}} & \multicolumn{1}{c}{\textbf{CIDEr}} \\
\midrule
\multirow[c]{4}{*}{\hspace{0.5em}32} & Ours & 57.30 $\pm$ 0.84 & 40.73 $\pm$ 1.12 & 52.79 $\pm$ 1.02 & 35.51 $\pm$ 1.16 & 53.88 $\pm$ 1.10 & 132.83 $\pm$ 4.72 \\
 & FT Projector & 52.60 $\pm$ 0.57 & 36.07 $\pm$ 0.67 & 49.02 $\pm$ 0.82 & 35.26 $\pm$ 1.61 & 49.54 $\pm$ 0.71 & 100.72 $\pm$ 7.59 \\
 & Projector & 53.81 $\pm$ 1.72 & 36.89 $\pm$ 2.49 & 50.37 $\pm$ 1.98 & 35.47 $\pm$ 3.52 & 49.73 $\pm$ 1.94 & 122.45 $\pm$ 12.01 \\
 & LoRA & 48.46 $\pm$ 0.37 & 29.15 $\pm$ 0.58 & 44.91 $\pm$ 0.42 & 24.71 $\pm$ 1.72 & 44.52 $\pm$ 0.87 & 62.11 $\pm$ 8.35\vspace{0.75em}\\
\multirow[c]{4}{*}{\hspace{0.5em}128} & Ours & 62.18 $\pm$ 0.54 & 46.00 $\pm$ 0.75 & 59.42 $\pm$ 0.76 & 44.13 $\pm$ 0.92 & 59.04 $\pm$ 0.65 & 171.15 $\pm$ 4.62 \\
 & FT Projector & 57.77 $\pm$ 0.66 & 41.45 $\pm$ 0.95 & 54.31 $\pm$ 0.73 & 40.07 $\pm$ 0.57 & 54.35 $\pm$ 0.57 & 147.71 $\pm$ 2.34 \\
 & Projector & 60.67 $\pm$ 0.97 & 44.44 $\pm$ 0.97 & 57.23 $\pm$ 0.84 & 41.02 $\pm$ 1.12 & 57.91 $\pm$ 1.29 & 158.41 $\pm$ 5.59 \\
\bottomrule
\end{tabular}}
\end{table}

\begin{table}[!htbp]
    \centering
    \caption{All results and metrics for the RN-50 encoder on the SydneyCaptions dataset. We show the mean ± standard error calculated over three random seeds.}
\resizebox{\textwidth}{!}{\begin{tabular}{llrrrrrr}
\toprule
\textbf{Sample Size} & \textbf{Setup} \textbackslash{} \textbf{Metric} & \multicolumn{1}{c}{\textbf{ROUGE-1}} & \multicolumn{1}{c}{\textbf{ROUGE-2}} & \multicolumn{1}{c}{\textbf{ROUGE-L}} & \multicolumn{1}{c}{\textbf{BLEU}} & \multicolumn{1}{c}{\textbf{METEOR}} & \multicolumn{1}{c}{\textbf{CIDEr}} \\
\midrule
\multirow[c]{4}{*}{\hspace{0.5em}32} & Ours & 59.49 $\pm$ 0.85 & 44.70 $\pm$ 1.23 & 56.91 $\pm$ 1.03 & 43.88 $\pm$ 1.27 & 55.15 $\pm$ 0.87 & 165.44 $\pm$ 2.06 \\
 & FT Projector & 55.53 $\pm$ 0.61 & 39.41 $\pm$ 1.14 & 52.75 $\pm$ 0.68 & 39.18 $\pm$ 1.31 & 51.58 $\pm$ 0.79 & 140.46 $\pm$ 6.76 \\
 & Projector & 50.68 $\pm$ 1.95 & 33.06 $\pm$ 2.52 & 47.45 $\pm$ 2.16 & 31.66 $\pm$ 2.99 & 46.20 $\pm$ 1.71 & 94.34 $\pm$ 16.68 \\
 & LoRA & 45.65 $\pm$ 0.43 & 27.34 $\pm$ 0.54 & 42.76 $\pm$ 0.50 & 24.72 $\pm$ 0.89 & 41.30 $\pm$ 0.56 & 70.76 $\pm$ 5.48\vspace{0.75em}\\
\multirow[c]{4}{*}{\hspace{0.5em}128} & Ours & 60.99 $\pm$ 0.16 & 44.20 $\pm$ 0.21 & 58.02 $\pm$ 0.18 & 41.87 $\pm$ 1.00 & 58.93 $\pm$ 0.35 & 157.43 $\pm$ 2.32 \\
 & FT Projector & 58.23 $\pm$ 0.53 & 39.79 $\pm$ 0.49 & 54.60 $\pm$ 0.57 & 37.55 $\pm$ 0.45 & 54.38 $\pm$ 0.56 & 128.26 $\pm$ 3.65 \\
 & Projector & 56.04 $\pm$ 1.74 & 36.95 $\pm$ 1.72 & 52.05 $\pm$ 1.67 & 33.55 $\pm$ 2.67 & 53.18 $\pm$ 2.06 & 112.26 $\pm$ 11.31 \\
\bottomrule
\end{tabular}}
\end{table}

\newpage

\subsubsection{CAPDELS dataset} \label{sec:add_mtrc_cap_8b}

\begin{table}[!htbp]
    \centering
    \caption{All results and metrics for the ConvNeXt-Nano encoder on the CAPDELS dataset.  We show the mean ± standard error calculated over three random seeds.}
\resizebox{\textwidth}{!}{\begin{tabular}{llrrrrrr}
\toprule
\textbf{Sample Size} & \textbf{Setup} \textbackslash{} \textbf{Metric} & \multicolumn{1}{c}{\textbf{ROUGE-1}} & \multicolumn{1}{c}{\textbf{ROUGE-2}} & \multicolumn{1}{c}{\textbf{ROUGE-L}} & \multicolumn{1}{c}{\textbf{BLEU}} & \multicolumn{1}{c}{\textbf{METEOR}} & \multicolumn{1}{c}{\textbf{CIDEr}} \\
\midrule
\multirow[c]{4}{*}{\hspace{0.5em}32} & Ours & 66.19 $\pm$ 0.41 & 49.52 $\pm$ 0.66 & 58.98 $\pm$ 0.66 & 46.03 $\pm$ 0.75 & 62.10 $\pm$ 0.54 & 215.76 $\pm$ 4.61 \\
 & FT Projector & 56.45 $\pm$ 0.51 & 36.95 $\pm$ 0.86 & 48.19 $\pm$ 0.63 & 31.84 $\pm$ 1.08 & 49.90 $\pm$ 0.43 & 104.63 $\pm$ 7.48 \\
 & Projector & 49.71 $\pm$ 1.96 & 31.01 $\pm$ 2.23 & 42.77 $\pm$ 2.04 & 26.51 $\pm$ 2.22 & 43.23 $\pm$ 1.53 & 87.04 $\pm$ 12.03 \\
 & LoRA & 46.75 $\pm$ 1.11 & 24.89 $\pm$ 2.02 & 38.18 $\pm$ 1.50 & 19.73 $\pm$ 1.57 & 38.61 $\pm$ 1.46 & 30.26 $\pm$ 5.83\vspace{0.75em}\\
\multirow[c]{4}{*}{\hspace{0.5em}128} & Ours & 69.71 $\pm$ 0.24 & 53.28 $\pm$ 0.54 & 62.26 $\pm$ 0.36 & 48.45 $\pm$ 0.48 & 66.57 $\pm$ 0.20 & 247.60 $\pm$ 10.07 \\
 & FT Projector & 63.37 $\pm$ 0.58 & 45.29 $\pm$ 0.89 & 55.28 $\pm$ 0.78 & 39.17 $\pm$ 1.02 & 59.16 $\pm$ 0.29 & 188.85 $\pm$ 6.96 \\
 & Projector & 60.76 $\pm$ 0.68 & 42.76 $\pm$ 0.74 & 53.31 $\pm$ 0.77 & 38.46 $\pm$ 1.25 & 55.15 $\pm$ 0.64 & 168.02 $\pm$ 5.11 \\
\bottomrule
\end{tabular}}
\end{table}

\begin{table}[!htbp]
    \centering
    \caption{All results and metrics for the ConvNeXt-Tiny encoder on the CAPDELS dataset.  We show the mean ± standard error calculated over three random seeds.}
\resizebox{\textwidth}{!}{\begin{tabular}{llrrrrrr}
\toprule
\textbf{Sample Size} & \textbf{Setup} \textbackslash{} \textbf{Metric} & \multicolumn{1}{c}{\textbf{ROUGE-1}} & \multicolumn{1}{c}{\textbf{ROUGE-2}} & \multicolumn{1}{c}{\textbf{ROUGE-L}} & \multicolumn{1}{c}{\textbf{BLEU}} & \multicolumn{1}{c}{\textbf{METEOR}} & \multicolumn{1}{c}{\textbf{CIDEr}} \\
\midrule
\multirow[c]{4}{*}{\hspace{0.5em}32} & Ours & 66.79 $\pm$ 0.46 & 50.40 $\pm$ 0.96 & 59.80 $\pm$ 0.75 & 46.01 $\pm$ 0.58 & 62.54 $\pm$ 0.54 & 219.97 $\pm$ 5.52 \\
 & FT Projector & 56.39 $\pm$ 0.53 & 37.45 $\pm$ 0.37 & 48.65 $\pm$ 0.41 & 32.92 $\pm$ 0.56 & 50.26 $\pm$ 0.74 & 94.00 $\pm$ 5.16 \\
 & Projector & 47.00 $\pm$ 2.81 & 28.43 $\pm$ 3.11 & 40.38 $\pm$ 2.78 & 23.95 $\pm$ 2.64 & 39.34 $\pm$ 2.90 & 61.45 $\pm$ 8.38 \\
 & LoRA & 43.96 $\pm$ 0.72 & 22.07 $\pm$ 1.07 & 35.35 $\pm$ 0.90 & 17.26 $\pm$ 1.10 & 36.61 $\pm$ 0.46 & 14.74 $\pm$ 3.98\vspace{0.75em}\\
\multirow[c]{4}{*}{\hspace{0.5em}128} & Ours & 71.48 $\pm$ 0.36 & 56.29 $\pm$ 0.44 & 64.86 $\pm$ 0.36 & 52.72 $\pm$ 0.69 & 68.36 $\pm$ 0.35 & 267.19 $\pm$ 9.72 \\
 & FT Projector & 63.87 $\pm$ 0.58 & 46.16 $\pm$ 0.65 & 56.08 $\pm$ 0.62 & 39.04 $\pm$ 1.12 & 59.78 $\pm$ 0.77 & 186.83 $\pm$ 1.87 \\
 & Projector & 61.92 $\pm$ 1.18 & 44.59 $\pm$ 1.61 & 54.61 $\pm$ 1.31 & 39.86 $\pm$ 1.83 & 56.58 $\pm$ 1.24 & 182.55 $\pm$ 15.27 \\
\bottomrule
\end{tabular}}
\end{table}

\begin{table}[!htbp]
    \centering
    \caption{All results and metrics for the ConvNeXt-Base encoder on the CAPDELS dataset.  We show the mean ± standard error calculated over three random seeds.}
\resizebox{\textwidth}{!}{\begin{tabular}{llrrrrrr}
\toprule
\textbf{Sample Size} & \textbf{Setup} \textbackslash{} \textbf{Metric} & \multicolumn{1}{c}{\textbf{ROUGE-1}} & \multicolumn{1}{c}{\textbf{ROUGE-2}} & \multicolumn{1}{c}{\textbf{ROUGE-L}} & \multicolumn{1}{c}{\textbf{BLEU}} & \multicolumn{1}{c}{\textbf{METEOR}} & \multicolumn{1}{c}{\textbf{CIDEr}} \\
\midrule
\multirow[c]{4}{*}{\hspace{0.5em}32} & Ours & 66.98 $\pm$ 0.34 & 51.12 $\pm$ 0.19 & 60.36 $\pm$ 0.34 & 46.52 $\pm$ 0.26 & 62.25 $\pm$ 0.55 & 231.03 $\pm$ 4.55 \\
 & FT Projector & 55.77 $\pm$ 0.33 & 36.49 $\pm$ 0.49 & 47.63 $\pm$ 0.38 & 31.26 $\pm$ 0.75 & 49.68 $\pm$ 0.38 & 105.70 $\pm$ 7.56 \\
 & Projector & 51.34 $\pm$ 1.49 & 32.48 $\pm$ 1.89 & 44.20 $\pm$ 1.54 & 26.80 $\pm$ 1.47 & 43.92 $\pm$ 1.70 & 89.78 $\pm$ 7.57 \\
 & LoRA & 43.32 $\pm$ 0.28 & 22.49 $\pm$ 0.56 & 34.69 $\pm$ 0.49 & 18.48 $\pm$ 0.47 & 36.59 $\pm$ 0.21 & 16.30 $\pm$ 6.21\vspace{0.75em}\\
\multirow[c]{4}{*}{\hspace{0.5em}128} & Ours & 72.14 $\pm$ 0.32 & 57.04 $\pm$ 0.44 & 65.55 $\pm$ 0.37 & 53.57 $\pm$ 0.79 & 68.84 $\pm$ 0.36 & 274.00 $\pm$ 2.80 \\
 & FT Projector & 65.35 $\pm$ 0.36 & 48.13 $\pm$ 0.50 & 57.67 $\pm$ 0.47 & 41.58 $\pm$ 0.63 & 61.45 $\pm$ 0.25 & 198.49 $\pm$ 7.28 \\
 & Projector & 61.94 $\pm$ 0.21 & 44.63 $\pm$ 0.23 & 54.61 $\pm$ 0.18 & 39.35 $\pm$ 0.66 & 57.39 $\pm$ 0.57 & 191.78 $\pm$ 1.90 \\
\bottomrule
\end{tabular}}
\end{table}

\newpage

\subsubsection{SensorCaps dataset} \label{sec:add_mtrc_sc_8b}

\begin{table}[!htbp]
    \centering
    \caption{All results and metrics for the LIMU-BERT encoder on the SensorCaps dataset. We show the mean ± standard error calculated over three random seeds.}
\resizebox{\textwidth}{!}{\begin{tabular}{llrrrrr}
\toprule
\textbf{Sample Size} & \textbf{Setup} \textbackslash{} \textbf{Metric} & \multicolumn{1}{c}{\textbf{ROUGE-1}} & \multicolumn{1}{c}{\textbf{ROUGE-2}} & \multicolumn{1}{c}{\textbf{ROUGE-L}} & \multicolumn{1}{c}{\textbf{BLEU}} & \multicolumn{1}{c}{\textbf{METEOR}} \\
\midrule
\multirow[c]{4}{*}{\hspace{0.5em}32} & Ours & 38.92 $\pm$ 0.55 & 11.61 $\pm$ 0.50 & 23.11 $\pm$ 0.80 & 8.46 $\pm$ 0.71 & 29.78 $\pm$ 0.82 \\
 & FT Projector & 34.80 $\pm$ 3.12 & 9.95 $\pm$ 1.13 & 21.35 $\pm$ 1.65 & 7.15 $\pm$ 1.11 & 26.91 $\pm$ 2.88 \\
 & Projector & 23.39 $\pm$ 1.95 & 4.69 $\pm$ 1.35 & 14.99 $\pm$ 1.03 & 3.09 $\pm$ 0.37 & 17.16 $\pm$ 0.68 \\
 & LoRA & 32.92 $\pm$ 0.33 & 8.97 $\pm$ 0.26 & 19.71 $\pm$ 0.26 & 6.18 $\pm$ 0.04 & 25.02 $\pm$ 0.35\vspace{0.75em}\\
\multirow[c]{4}{*}{\hspace{0.5em}128} & Ours & 48.36 $\pm$ 0.83 & 18.54 $\pm$ 0.71 & 30.63 $\pm$ 0.61 & 15.66 $\pm$ 0.62 & 39.49 $\pm$ 0.83 \\
 & FT Projector & 44.98 $\pm$ 0.78 & 15.48 $\pm$ 0.35 & 27.56 $\pm$ 0.43 & 12.71 $\pm$ 0.48 & 35.80 $\pm$ 0.72 \\
 & Projector & 35.31 $\pm$ 4.27 & 10.85 $\pm$ 2.55 & 21.68 $\pm$ 2.97 & 8.18 $\pm$ 2.19 & 26.78 $\pm$ 3.90 \\
\bottomrule
\end{tabular}}
\end{table}

\newpage

\subsubsection{ChEBI-20 dataset} \label{sec:add_mtrc_che_8b}

\begin{table}[!htbp]
    \centering
    \caption{All results and metrics for the MolCA encoder on the ChEBI-20 dataset. We show the single seed results.}
\begin{tabular}{llrrrrr}
\toprule
\textbf{Sample Size} & \textbf{Setup} \textbackslash{} \textbf{Metric} & \multicolumn{1}{c}{\textbf{ROUGE-1}} & \multicolumn{1}{c}{\textbf{ROUGE-2}} & \multicolumn{1}{c}{\textbf{ROUGE-L}} & \multicolumn{1}{c}{\textbf{BLEU}} & \multicolumn{1}{c}{\textbf{METEOR}} \\
\midrule
\multirow[c]{4}{*}{\hspace{0.5em}32} & Ours & 32.30 & 13.67 & 26.64 & 8.93 & 20.62 \\
 & FT Projector & 21.27 & 6.52 & 19.42 & 0.75 & 7.32 \\
 & Projector & 20.97 & 6.68 & 15.97 & 3.59 & 13.93 \\
 & LoRA & 22.26 & 7.38 & 18.30 & 3.26 & 12.81\vspace{0.75em}\\
\multirow[c]{4}{*}{\hspace{0.5em}128} & Ours & 33.35 & 14.05 & 26.99 & 9.31 & 22.20 \\
 & FT Projector & 26.66 & 11.11 & 22.11 & 5.89 & 16.73 \\
 & Projector & 18.93 & 6.13 & 15.09 & 4.13 & 12.81 \\
 & LoRA & 30.11 & 12.29 & 24.23 & 7.36 & 19.90\vspace{0.75em}\\
\multirow[c]{4}{*}{\hspace{0.5em}512} & Ours & 35.49 & 17.61 & 30.26 & 11.91 & 23.85 \\
 & FT Projector & 32.59 & 14.44 & 26.33 & 9.68 & 20.87 \\
 & Projector & 24.88 & 10.17 & 20.46 & 5.84 & 14.72 \\
 & LoRA & 32.43 & 14.13 & 26.37 & 9.46 & 20.67\vspace{0.75em}\\
\multirow[c]{4}{*}{\hspace{0.5em}2048} & Ours & 39.91 & 20.60 & 33.50 & 14.22 & 29.44 \\
 & FT Projector & 37.88 & 18.17 & 30.65 & 13.36 & 26.47 \\
 & Projector & 35.37 & 16.49 & 29.29 & 10.87 & 23.71 \\
 & LoRA & 34.83 & 15.88 & 28.38 & 10.86 & 25.41\vspace{0.75em}\\
\multirow[c]{4}{*}{\hspace{0.5em}8192} & Ours & 42.67 & 23.57 & 36.69 & 17.19 & 31.01 \\
 & FT Projector & 43.77 & 23.33 & 36.17 & 19.30 & 33.32 \\
 & Projector & 32.21 & 14.34 & 26.25 & 8.42 & 22.73 \\
 & LoRA & 40.17 & 21.43 & 34.44 & 15.30 & 29.01\vspace{0.75em}\\
\multirow[c]{4}{*}{\hspace{0.5em}226407} & Ours & 44.55 & 24.65 & 37.41 & 18.61 & 35.64 \\
 & FT Projector & 47.24 & 26.93 & 39.53 & 22.57 & 37.40 \\
 & Projector & 36.54 & 17.94 & 30.26 & 11.49 & 26.55 \\
 & LoRA & 40.00 & 21.33 & 34.20 & 15.05 & 28.60\\
\bottomrule
\end{tabular}
\end{table}

\newpage

\subsubsection{SoundBible dataset} \label{sec:add_mtrc_sb_8b}

\begin{table}[!htbp]
    \centering
    \caption{All results and metrics for the BLAT encoder on the SoundBible dataset. We show the mean ± standard error calculated over three random seeds.}
\resizebox{\textwidth}{!}{\begin{tabular}{lllllll}
\toprule
\textbf{Sample Size} & \textbf{Setup} \textbackslash{} \textbf{Metric} & \multicolumn{1}{c}{\textbf{ROUGE-1}} & \multicolumn{1}{c}{\textbf{ROUGE-2}} & \multicolumn{1}{c}{\textbf{ROUGE-L}} & \multicolumn{1}{c}{\textbf{BLEU}} & \multicolumn{1}{c}{\textbf{METEOR}} \\
\midrule
\multirow[c]{4}{*}{\hspace{0.5em}32} & Ours & 27.20 $\pm$ 0.44 & 3.52 $\pm$ 0.14 & 25.49 $\pm$ 0.49 & 3.39 $\pm$ 0.10 & 23.03 $\pm$ 0.16 \\
 & FT Projector & 27.42 $\pm$ 0.10 & 4.51 $\pm$ 0.07 & 25.36 $\pm$ 0.15 & 3.36 $\pm$ 0.02 & 24.15 $\pm$ 0.11 \\
 & Projector & 24.60 $\pm$ 1.24 & 2.87 $\pm$ 0.11 & 22.95 $\pm$ 1.14 & 2.75 $\pm$ 0.19 & 21.76 $\pm$ 0.70 \\
 & LoRA & 25.69 $\pm$ 0.84 & 2.50 $\pm$ 0.27 & 23.59 $\pm$ 0.83 & 2.69 $\pm$ 0.18 & 22.13 $\pm$ 0.59\vspace{0.75em}\\
\multirow[c]{4}{*}{\hspace{0.5em}128} & Ours & 30.32 $\pm$ 0.47 & 8.39 $\pm$ 0.36 & 28.96 $\pm$ 0.44 & 6.87 $\pm$ 0.21 & 26.74 $\pm$ 0.31 \\
 & FT Projector & 31.52 $\pm$ 0.40 & 8.94 $\pm$ 0.37 & 29.82 $\pm$ 0.34 & 7.29 $\pm$ 0.35 & 27.92 $\pm$ 0.45 \\
 & Projector & 29.74 $\pm$ 0.71 & 8.74 $\pm$ 0.25 & 28.02 $\pm$ 0.63 & 6.92 $\pm$ 0.22 & 26.88 $\pm$ 0.43 \\
\bottomrule
\end{tabular}}
\end{table}

\newpage

\subsection{Llama 3.2 1B Instruct Results}
\label{sec:add_metrics_1b}

\subsubsection{SydneyCaptions dataset} \label{sec:add_mtrc_syd_1b}

\begin{table}[!htbp]
    \centering
    \caption{All results and metrics for the ViT-B-32 encoder on the SydneyCaptions dataset. We show the mean ± standard error calculated over five random seeds.}
\resizebox{\textwidth}{!}{\begin{tabular}{llrrrrrr}
\toprule
\textbf{Sample Size} & \textbf{Setup} \textbackslash{} \textbf{Metric} & \multicolumn{1}{c}{\textbf{ROUGE-1}} & \multicolumn{1}{c}{\textbf{ROUGE-2}} & \multicolumn{1}{c}{\textbf{ROUGE-L}} & \multicolumn{1}{c}{\textbf{BLEU}} & \multicolumn{1}{c}{\textbf{METEOR}} & \multicolumn{1}{c}{\textbf{CIDEr}} \\

\midrule
\multirow[c]{4}{*}{\hspace{0.5em}32} & Ours & 59.40 $\pm$ 2.15 & 43.82 $\pm$ 2.16 & 57.06 $\pm$ 2.16 & 41.34 $\pm$ 2.23 & 55.44 $\pm$ 2.46 & 166.58 $\pm$ 7.35 \\
 & FT Projector & 48.91 $\pm$ 0.17 & 30.12 $\pm$ 0.55 & 45.48 $\pm$ 0.31 & 28.16 $\pm$ 0.60 & 42.32 $\pm$ 0.38 & 121.49 $\pm$ 4.14 \\
 & Projector & 47.67 $\pm$ 1.02 & 30.54 $\pm$ 1.41 & 44.81 $\pm$ 1.06 & 29.82 $\pm$ 2.78 & 42.32 $\pm$ 0.98 & 128.17 $\pm$ 5.10 \\
 & LoRA & 47.91 $\pm$ 0.51 & 29.65 $\pm$ 0.96 & 44.86 $\pm$ 0.63 & 26.65 $\pm$ 1.04 & 41.92 $\pm$ 0.64 & 109.91 $\pm$ 7.97\vspace{0.75em}\\
\multirow[c]{4}{*}{\hspace{0.5em}128} & Ours & 59.97 $\pm$ 2.07 & 43.80 $\pm$ 2.54 & 57.14 $\pm$ 1.96 & 37.92 $\pm$ 3.01 & 58.65 $\pm$ 1.92 & 148.99 $\pm$ 17.39 \\
 & FT Projector & 60.15 $\pm$ 1.30 & 42.36 $\pm$ 1.65 & 56.87 $\pm$ 1.46 & 39.44 $\pm$ 1.71 & 55.65 $\pm$ 1.60 & 157.42 $\pm$ 7.03 \\
 & Projector & 58.21 $\pm$ 1.20 & 42.07 $\pm$ 1.52 & 55.15 $\pm$ 1.43 & 38.47 $\pm$ 0.97 & 54.92 $\pm$ 1.62 & 151.02 $\pm$ 2.11 \\
\bottomrule
\end{tabular}}
\end{table}

\begin{table}[!htbp]
    \centering
    \caption{All results and metrics for the ViT-L-14 encoder on the SydneyCaptions dataset. We show the mean ± standard error calculated over five random seeds.}
\resizebox{\textwidth}{!}{\begin{tabular}{llrrrrrr}
\toprule
\textbf{Sample Size} & \textbf{Setup} \textbackslash{} \textbf{Metric} & \multicolumn{1}{c}{\textbf{ROUGE-1}} & \multicolumn{1}{c}{\textbf{ROUGE-2}} & \multicolumn{1}{c}{\textbf{ROUGE-L}} & \multicolumn{1}{c}{\textbf{BLEU}} & \multicolumn{1}{c}{\textbf{METEOR}} & \multicolumn{1}{c}{\textbf{CIDEr}} \\
\midrule
\multirow[c]{4}{*}{\hspace{0.5em}32} & Ours & 57.95 $\pm$ 0.83 & 41.15 $\pm$ 1.11 & 54.02 $\pm$ 0.80 & 36.37 $\pm$ 0.75 & 55.06 $\pm$ 1.02 & 133.65 $\pm$ 2.69 \\
 & FT Projector & 51.19 $\pm$ 1.43 & 35.00 $\pm$ 1.49 & 48.36 $\pm$ 1.54 & 31.06 $\pm$ 1.73 & 45.31 $\pm$ 1.42 & 80.49 $\pm$ 5.57 \\
 & Projector & 54.31 $\pm$ 1.76 & 38.65 $\pm$ 2.21 & 51.06 $\pm$ 1.54 & 35.90 $\pm$ 2.69 & 50.71 $\pm$ 2.22 & 113.79 $\pm$ 6.87 \\
 & LoRA & 47.93 $\pm$ 1.52 & 31.50 $\pm$ 1.93 & 45.40 $\pm$ 1.65 & 28.02 $\pm$ 2.96 & 42.03 $\pm$ 1.23 & 69.21 $\pm$ 7.36\vspace{0.75em}\\
\multirow[c]{4}{*}{\hspace{0.5em}128} & Ours & 62.65 $\pm$ 0.64 & 47.53 $\pm$ 1.16 & 59.69 $\pm$ 0.83 & 45.17 $\pm$ 1.84 & 58.92 $\pm$ 0.62 & 173.37 $\pm$ 4.04 \\
 & FT Projector & 55.87 $\pm$ 0.71 & 39.14 $\pm$ 0.94 & 53.30 $\pm$ 0.80 & 31.92 $\pm$ 1.16 & 52.09 $\pm$ 0.58 & 122.39 $\pm$ 2.46 \\
 & Projector & 53.68 $\pm$ 2.45 & 38.27 $\pm$ 2.53 & 51.07 $\pm$ 2.56 & 34.69 $\pm$ 3.45 & 50.15 $\pm$ 2.31 & 133.22 $\pm$ 9.80 \\
\bottomrule
\end{tabular}}
\end{table}

\begin{table}[!htbp]
    \centering
    \caption{All results and metrics for the RN-50 encoder on the SydneyCaptions dataset. We show the mean ± standard error calculated over five random seeds.}
\resizebox{\textwidth}{!}{\begin{tabular}{llrrrrrr}
\toprule
\textbf{Sample Size} & \textbf{Setup} \textbackslash{} \textbf{Metric} & \multicolumn{1}{c}{\textbf{ROUGE-1}} & \multicolumn{1}{c}{\textbf{ROUGE-2}} & \multicolumn{1}{c}{\textbf{ROUGE-L}} & \multicolumn{1}{c}{\textbf{BLEU}} & \multicolumn{1}{c}{\textbf{METEOR}} & \multicolumn{1}{c}{\textbf{CIDEr}} \\

\midrule
\multirow[c]{4}{*}{\hspace{0.5em}32} & Ours & 58.05 $\pm$ 0.33 & 42.76 $\pm$ 1.03 & 55.17 $\pm$ 0.68 & 42.34 $\pm$ 1.37 & 53.45 $\pm$ 0.52 & 153.84 $\pm$ 5.53 \\
 & FT Projector & 56.75 $\pm$ 0.85 & 39.59 $\pm$ 0.85 & 53.35 $\pm$ 0.70 & 38.67 $\pm$ 1.36 & 51.79 $\pm$ 0.95 & 130.78 $\pm$ 4.67 \\
 & Projector & 51.09 $\pm$ 3.35 & 33.71 $\pm$ 4.34 & 47.76 $\pm$ 3.53 & 29.97 $\pm$ 5.36 & 46.67 $\pm$ 3.58 & 107.43 $\pm$ 22.35 \\
 & LoRA & 47.91 $\pm$ 0.52 & 26.35 $\pm$ 1.43 & 44.63 $\pm$ 0.84 & 23.36 $\pm$ 1.86 & 41.03 $\pm$ 0.86 & 61.69 $\pm$ 1.26\vspace{0.75em}\\
\multirow[c]{4}{*}{\hspace{0.5em}128} & Ours & 57.78 $\pm$ 1.70 & 39.78 $\pm$ 2.26 & 54.37 $\pm$ 2.24 & 35.94 $\pm$ 3.07 & 55.09 $\pm$ 2.01 & 128.53 $\pm$ 10.50 \\
 & FT Projector & 49.10 $\pm$ 0.60 & 29.71 $\pm$ 0.96 & 45.28 $\pm$ 0.41 & 28.60 $\pm$ 1.50 & 44.30 $\pm$ 0.56 & 67.96 $\pm$ 4.25 \\
 & Projector & 50.07 $\pm$ 2.25 & 30.65 $\pm$ 2.90 & 46.19 $\pm$ 2.37 & 26.83 $\pm$ 2.98 & 45.41 $\pm$ 2.46 & 85.86 $\pm$ 14.36 \\
\bottomrule
\end{tabular}}
\end{table}

\newpage

\subsubsection{CAPDELS dataset} \label{sec:add_mtrc_cap_1b}

\begin{table}[!htbp]
    \centering
    \caption{All results and metrics for the ConvNeXt-Nano encoder on the CAPDELS dataset.  We show the mean ± standard error calculated over five random seeds.}
\resizebox{\textwidth}{!}{\begin{tabular}{llrrrrrr}
\toprule
\textbf{Sample Size} & \textbf{Setup} \textbackslash{} \textbf{Metric} & \multicolumn{1}{c}{\textbf{ROUGE-1}} & \multicolumn{1}{c}{\textbf{ROUGE-2}} & \multicolumn{1}{c}{\textbf{ROUGE-L}} & \multicolumn{1}{c}{\textbf{BLEU}} & \multicolumn{1}{c}{\textbf{METEOR}} & \multicolumn{1}{c}{\textbf{CIDEr}} \\

\midrule
\multirow[c]{4}{*}{\hspace{0.5em}32} & Ours & 52.08 $\pm$ 1.22 & 35.44 $\pm$ 1.20 & 45.60 $\pm$ 1.12 & 30.29 $\pm$ 0.91 & 46.07 $\pm$ 1.37 & 135.57 $\pm$ 9.21 \\
 & FT Projector & 43.76 $\pm$ 0.59 & 24.47 $\pm$ 0.62 & 36.21 $\pm$ 0.62 & 19.16 $\pm$ 0.48 & 39.47 $\pm$ 0.81 & 57.40 $\pm$ 1.52 \\
 & Projector & 37.99 $\pm$ 2.35 & 20.59 $\pm$ 1.91 & 31.68 $\pm$ 2.09 & 15.49 $\pm$ 1.89 & 32.46 $\pm$ 1.74 & 34.80 $\pm$ 4.73 \\
 & LoRA & 36.61 $\pm$ 1.51 & 18.49 $\pm$ 1.76 & 29.70 $\pm$ 1.66 & 14.62 $\pm$ 2.00 & 33.68 $\pm$ 1.16 & 27.40 $\pm$ 4.31\vspace{0.75em}\\
\multirow[c]{4}{*}{\hspace{0.5em}128} & Ours & 60.34 $\pm$ 0.58 & 43.06 $\pm$ 0.64 & 53.13 $\pm$ 0.58 & 37.89 $\pm$ 0.89 & 55.70 $\pm$ 0.53 & 189.89 $\pm$ 7.42 \\
 & FT Projector & 53.89 $\pm$ 0.60 & 35.25 $\pm$ 0.24 & 45.96 $\pm$ 0.31 & 28.14 $\pm$ 0.25 & 49.38 $\pm$ 0.71 & 135.04 $\pm$ 2.15 \\
 & Projector & 51.95 $\pm$ 1.63 & 33.71 $\pm$ 1.45 & 44.33 $\pm$ 1.63 & 28.13 $\pm$ 1.82 & 47.88 $\pm$ 1.46 & 124.49 $\pm$ 10.98 \\
\bottomrule
\end{tabular}}
\end{table}

\begin{table}[!htbp]
    \centering
    \caption{All results and metrics for the ConvNeXt-Tiny encoder on the CAPDELS dataset.  We show the mean ± standard error calculated over five random seeds.}
\resizebox{\textwidth}{!}{\begin{tabular}{llrrrrrr}
\toprule
\textbf{Sample Size} & \textbf{Setup} \textbackslash{} \textbf{Metric} & \multicolumn{1}{c}{\textbf{ROUGE-1}} & \multicolumn{1}{c}{\textbf{ROUGE-2}} & \multicolumn{1}{c}{\textbf{ROUGE-L}} & \multicolumn{1}{c}{\textbf{BLEU}} & \multicolumn{1}{c}{\textbf{METEOR}} & \multicolumn{1}{c}{\textbf{CIDEr}} \\

\midrule
\multirow[c]{4}{*}{\hspace{0.5em}32} & Ours & 54.04 $\pm$ 1.08 & 36.09 $\pm$ 1.12 & 46.90 $\pm$ 1.02 & 31.13 $\pm$ 1.48 & 48.41 $\pm$ 1.44 & 136.77 $\pm$ 13.36 \\
 & FT Projector & 45.23 $\pm$ 0.73 & 26.56 $\pm$ 0.92 & 37.83 $\pm$ 0.77 & 21.86 $\pm$ 0.81 & 39.61 $\pm$ 0.53 & 70.52 $\pm$ 4.90 \\
 & Projector & 35.84 $\pm$ 3.52 & 17.22 $\pm$ 3.40 & 28.97 $\pm$ 3.25 & 13.61 $\pm$ 2.76 & 30.71 $\pm$ 3.79 & 34.94 $\pm$ 11.34 \\
 & LoRA & 41.07 $\pm$ 1.18 & 21.35 $\pm$ 1.50 & 33.26 $\pm$ 1.31 & 17.53 $\pm$ 1.42 & 35.01 $\pm$ 1.22 & 33.87 $\pm$ 6.49\vspace{0.75em}\\
\multirow[c]{4}{*}{\hspace{0.5em}128} & Ours & 61.18 $\pm$ 1.06 & 44.13 $\pm$ 1.44 & 53.93 $\pm$ 1.22 & 39.95 $\pm$ 1.52 & 56.25 $\pm$ 1.21 & 178.43 $\pm$ 14.90 \\
 & FT Projector & 52.64 $\pm$ 0.51 & 35.05 $\pm$ 0.71 & 45.76 $\pm$ 0.63 & 30.25 $\pm$ 0.86 & 48.16 $\pm$ 0.45 & 118.66 $\pm$ 3.97 \\
 & Projector & 45.11 $\pm$ 5.37 & 26.98 $\pm$ 5.17 & 37.93 $\pm$ 5.13 & 21.92 $\pm$ 4.41 & 40.51 $\pm$ 5.37 & 77.51 $\pm$ 20.00 \\
\bottomrule
\end{tabular}}
\end{table}

\begin{table}[!htbp]
    \centering
    \caption{All results and metrics for the ConvNeXt-Base encoder on the CAPDELS dataset.  We show the mean ± standard error calculated over five random seeds.}
\resizebox{\textwidth}{!}{\begin{tabular}{llrrrrrr}
\toprule
\textbf{Sample Size} & \textbf{Setup} \textbackslash{} \textbf{Metric} & \multicolumn{1}{c}{\textbf{ROUGE-1}} & \multicolumn{1}{c}{\textbf{ROUGE-2}} & \multicolumn{1}{c}{\textbf{ROUGE-L}} & \multicolumn{1}{c}{\textbf{BLEU}} & \multicolumn{1}{c}{\textbf{METEOR}} & \multicolumn{1}{c}{\textbf{CIDEr}} \\

\midrule
\multirow[c]{4}{*}{\hspace{0.5em}32} & Ours & 50.67 $\pm$ 3.26 & 31.75 $\pm$ 4.45 & 43.22 $\pm$ 3.81 & 26.77 $\pm$ 3.84 & 44.57 $\pm$ 3.29 & 91.85 $\pm$ 27.84 \\
 & FT Projector & 45.45 $\pm$ 1.10 & 26.57 $\pm$ 1.38 & 38.17 $\pm$ 1.28 & 20.40 $\pm$ 1.39 & 40.94 $\pm$ 1.17 & 62.25 $\pm$ 5.29 \\
 & Projector & 37.00 $\pm$ 2.85 & 18.31 $\pm$ 3.01 & 30.20 $\pm$ 2.63 & 14.27 $\pm$ 2.59 & 31.47 $\pm$ 2.94 & 37.62 $\pm$ 10.70 \\
 & LoRA & 37.93 $\pm$ 0.80 & 18.07 $\pm$ 1.05 & 30.32 $\pm$ 0.95 & 14.13 $\pm$ 1.40 & 32.70 $\pm$ 0.69 & 18.05 $\pm$ 1.64\vspace{0.75em}\\
\multirow[c]{4}{*}{\hspace{0.5em}128} & Ours & 56.07 $\pm$ 3.56 & 39.14 $\pm$ 3.70 & 48.72 $\pm$ 3.95 & 35.83 $\pm$ 3.63 & 51.94 $\pm$ 3.35 & 151.69 $\pm$ 29.21 \\
 & FT Projector & 54.70 $\pm$ 0.64 & 36.64 $\pm$ 0.87 & 47.37 $\pm$ 0.74 & 30.75 $\pm$ 0.77 & 50.36 $\pm$ 0.66 & 132.67 $\pm$ 7.65 \\
 & Projector & 44.11 $\pm$ 3.46 & 25.02 $\pm$ 3.34 & 36.58 $\pm$ 3.20 & 20.51 $\pm$ 3.28 & 39.40 $\pm$ 3.37 & 67.66 $\pm$ 13.22 \\
\bottomrule
\end{tabular}}
\end{table}

\newpage
\subsubsection{SensorCaps dataset} \label{sec:add_mtrc_sc_1b}

\begin{table}[!htbp]
    \centering
    \caption{All results and metrics for the LIMU-BERT encoder on the SensorCaps dataset. We show the mean ± standard error calculated over five random seeds.}
\resizebox{\textwidth}{!}{\begin{tabular}{llrrrrr}
\toprule
\textbf{Sample Size} & \textbf{Setup} \textbackslash{} \textbf{Metric} & \multicolumn{1}{c}{\textbf{ROUGE-1}} & \multicolumn{1}{c}{\textbf{ROUGE-2}} & \multicolumn{1}{c}{\textbf{ROUGE-L}} & \multicolumn{1}{c}{\textbf{BLEU}} & \multicolumn{1}{c}{\textbf{METEOR}} \\
\midrule
\multirow[c]{4}{*}{\hspace{0.5em}32} & Ours & 27.87 $\pm$ 0.69 & 6.47 $\pm$ 0.37 & 17.09 $\pm$ 0.47 & 4.00 $\pm$ 0.28 & 20.11 $\pm$ 0.24 \\
 & FT Projector & 28.14 $\pm$ 0.36 & 5.86 $\pm$ 0.21 & 16.90 $\pm$ 0.22 & 3.48 $\pm$ 0.09 & 20.06 $\pm$ 0.33 \\
 & Projector & 22.06 $\pm$ 3.67 & 4.59 $\pm$ 1.20 & 14.28 $\pm$ 2.27 & 3.16 $\pm$ 0.91 & 16.52 $\pm$ 2.71 \\
 & LoRA & 30.26 $\pm$ 0.40 & 8.19 $\pm$ 0.31 & 17.88 $\pm$ 0.28 & 4.55 $\pm$ 0.11 & 21.53 $\pm$ 0.34\vspace{0.75em}\\
\multirow[c]{4}{*}{\hspace{0.5em}128} & Ours & 39.77 $\pm$ 1.11 & 12.93 $\pm$ 0.57 & 23.65 $\pm$ 0.63 & 9.02 $\pm$ 0.28 & 29.61 $\pm$ 0.62 \\
 & FT Projector & 36.10 $\pm$ 0.46 & 10.76 $\pm$ 0.29 & 21.69 $\pm$ 0.25 & 7.89 $\pm$ 0.29 & 26.79 $\pm$ 0.42 \\
 & Projector & 28.17 $\pm$ 4.72 & 7.53 $\pm$ 1.85 & 18.09 $\pm$ 2.91 & 5.21 $\pm$ 1.59 & 20.64 $\pm$ 3.92 \\
\bottomrule
\end{tabular}}
\end{table}

\subsubsection{ChEBI-20 dataset} \label{sec:add_mtrc_che_1b}

\begin{table}[!htbp]
    \centering
    \caption{All results and metrics for the MolCA encoder on the ChEBI-20 dataset. We show the mean ± standard error calculated over five random seeds.}
\resizebox{\textwidth}{!}{\begin{tabular}{llrrrrr}
\toprule
\textbf{Sample Size} & \textbf{Setup} \textbackslash{} \textbf{Metric} & \multicolumn{1}{c}{\textbf{ROUGE-1}} & \multicolumn{1}{c}{\textbf{ROUGE-2}} & \multicolumn{1}{c}{\textbf{ROUGE-L}} & \multicolumn{1}{c}{\textbf{BLEU}} & \multicolumn{1}{c}{\textbf{METEOR}} \\
\midrule
\multirow[c]{4}{*}{\hspace{0.5em}32} & Ours & 27.63 $\pm$ 0.97 & 10.80 $\pm$ 0.56 & 22.20 $\pm$ 0.80 & 6.26 $\pm$ 0.49 & 17.91 $\pm$ 0.57 \\
 & FT Projector & 18.56 $\pm$ 0.17 & 4.65 $\pm$ 0.08 & 14.72 $\pm$ 0.13 & 2.38 $\pm$ 0.05 & 9.89 $\pm$ 0.09 \\
 & Projector & 21.99 $\pm$ 1.06 & 8.40 $\pm$ 0.67 & 18.07 $\pm$ 1.01 & 4.19 $\pm$ 0.37 & 12.81 $\pm$ 0.62 \\
 & LoRA & 23.22 $\pm$ 0.40 & 6.82 $\pm$ 0.27 & 18.70 $\pm$ 0.31 & 3.83 $\pm$ 0.24 & 12.33 $\pm$ 0.44\vspace{0.75em}\\
\multirow[c]{4}{*}{\hspace{0.5em}128} & Ours & 29.60 $\pm$ 1.07 & 11.71 $\pm$ 1.26 & 23.65 $\pm$ 1.26 & 7.45 $\pm$ 0.85 & 17.86 $\pm$ 0.42 \\
 & FT Projector & 24.97 $\pm$ 0.17 & 7.90 $\pm$ 0.12 & 19.88 $\pm$ 0.15 & 4.70 $\pm$ 0.11 & 14.52 $\pm$ 0.15 \\
 & Projector & 25.27 $\pm$ 0.93 & 9.06 $\pm$ 1.01 & 20.47 $\pm$ 1.15 & 4.86 $\pm$ 0.52 & 14.68 $\pm$ 0.61 \\
\bottomrule
\end{tabular}}
\end{table}

\newpage

\subsubsection{SoundBible dataset} \label{sec:add_mtrc_sb_1b}

\begin{table}[!htbp]
    \centering
    \caption{All results and metrics for the BLAT encoder on the SoundBible dataset. We show the mean ± standard error calculated over five random seeds.}
\resizebox{\textwidth}{!}{\begin{tabular}{llrrrrr}
\toprule
\textbf{Sample Size} & \textbf{Setup} \textbackslash{} \textbf{Metric} & \multicolumn{1}{c}{\textbf{ROUGE-1}} & \multicolumn{1}{c}{\textbf{ROUGE-2}} & \multicolumn{1}{c}{\textbf{ROUGE-L}} & \multicolumn{1}{c}{\textbf{BLEU}} & \multicolumn{1}{c}{\textbf{METEOR}} \\
\midrule
\multirow[c]{4}{*}{\hspace{0.5em}32} & Ours & 26.83 $\pm$ 0.71 & 3.07 $\pm$ 0.22 & 25.28 $\pm$ 0.55 & 3.07 $\pm$ 0.18 & 22.55 $\pm$ 0.56 \\
 & FT Projector & 24.63 $\pm$ 0.35 & 3.82 $\pm$ 0.60 & 23.52 $\pm$ 0.31 & 3.33 $\pm$ 0.12 & 21.97 $\pm$ 0.41 \\
 & Projector & 20.50 $\pm$ 2.58 & 1.74 $\pm$ 0.48 & 18.94 $\pm$ 2.71 & 1.85 $\pm$ 0.56 & 18.30 $\pm$ 1.41 \\
 & LoRA & 20.67 $\pm$ 0.93 & 2.03 $\pm$ 0.33 & 19.10 $\pm$ 0.92 & 1.44 $\pm$ 0.60 & 18.95 $\pm$ 0.57\vspace{0.75em}\\
\multirow[c]{4}{*}{\hspace{0.5em}128} & Ours & 24.66 $\pm$ 0.72 & 5.29 $\pm$ 0.42 & 23.33 $\pm$ 0.73 & 4.10 $\pm$ 0.17 & 22.08 $\pm$ 0.58 \\
 & FT Projector & 24.51 $\pm$ 0.69 & 6.50 $\pm$ 0.33 & 22.83 $\pm$ 0.61 & 4.26 $\pm$ 0.28 & 23.43 $\pm$ 0.39 \\
 & Projector & 19.00 $\pm$ 1.59 & 2.75 $\pm$ 0.46 & 17.48 $\pm$ 1.65 & 1.58 $\pm$ 0.29 & 18.12 $\pm$ 0.90 \\
\bottomrule
\end{tabular}}
\end{table}

\section{Dataset Details} \label{sec:data_details}

The licences for datasets are provided in \cref{tab:licences}.

\begin{table}[htbp]
\centering
\begin{threeparttable} 
\caption{Dataset licences.} \label{tab:licences}
\centering
\begin{tabularx}{\textwidth}{@{}ll@{}}
\toprule
\textbf{Dataset} & \textbf{Licence} \\ 
\midrule
COCO \cite{lin2014microsoft} & Captions: CC BY 4.0; Images: Mixed (CC BY 4.0 \& other licences) \\
AudioCaps \cite{kim2019audiocaps} & Unknown \\
OpenVid \cite{nan2024openvid} & CC BY 4.0 \\
ShareGPT4V \cite{chen2023sharegpt4v} & CC BY-NC 4.0 \\
ShareGPT4Video \cite{chen2024sharegpt4video} & CC BY-NC 4.0 \\
Clotho \cite{bubogpt, clotho} & Captions: CC BY-NC; Audio: Mostly CC BY (version not provided) \\
SydneyCaptions \cite{sydneycaps} & Unknown \\
SensorCaps \cite{sensorcaps} & Hippocratic License\tnote{1}\\
CAPDELS (ours) & Derived from CANDELS dataset \cite{candels} - CC BY-NC-SA 4.0 \\
SoundBible \cite{wavcaps} & Custom licence\tnote{2} \\
\bottomrule 
\end{tabularx}
\begin{tablenotes}
\item[1] \href{https://firstdonoharm.dev/version/3/0/cl-eco-extr-ffd-law-media-mil-my-soc-sv-tal-usta.html}{Link for the licence}
\item[2] \href{https://soundbible.com/about.php}{SoundBible website}
\end{tablenotes} 
\end{threeparttable}
\end{table}

Unless otherwise specified, all datasets are utilised in their original form, without any alterations. Splits of low-resource modality datasets are adopted according to the specifications provided by each dataset’s authors.

\textbf{OpenVid}~~~We access the video zip files (with the template \texttt{OpenVid\_part\{index\}.zip}) through HuggingFace \cite{openvid_hf}. We randomly sample 9 zip files out of 185, specifically indices 1, 7, 15, 52, 80, 101, 125, 150, and 174. Moreover, only the first two sentences of the descriptions are used, determined by the \verb|sent_tokenize| function in the NLTK Python library \cite{nltk}.

\textbf{ShareGPT4V}~~~We access the dataset description files through HuggingFace \cite{sharegpt4v_hf}. Rather than using the all of the dataset, we only use the images sourced from LLaVA \cite{liu2023llava} and SegmentAnything \cite{segmentanything} datasets. Furthermore, if there are multiple descriptions for an image, we consider only one of them.

\textbf{ShareGPT4Video}~~~Rather than using full resolution videos, we use videos downscaled to 360P.

\textbf{SensorCaps}~~~We remove the final two sentences from the description, as they typically reiterate the preceding text and offer little additional information.

\textbf{ChEBI-20}~~~In addition to the molecule embedding extracted from the MolCA encoder, we augment the prompt with the SMILES string of the molecule.

\section{Training Details} \label{sec:train_details}

\subsection{Encoder Details}

In \cref{tab:encoders}, we list the checkpoint descriptions of encoders used in projector pre-training or hypernetwork training stages. These descriptions are either Hugging Face identifiers or specific checkpoint descriptions enabling reproduction of our results.

\begin{table}[!ht]
\centering
\caption{The checkpoint descriptions 
for each encoder used during training phases.} \label{tab:encoders}
\begin{tabular}{ll}
\toprule
\textbf{Encoder} & \textbf{Identifier} \\ \midrule
CLIP \cite{radford2021clip}   &  \verb|openai/clip-vit-large-patch14|  \\
CLAP \cite{elizalde2023clap} &  \verb|laion/clap-htsat-fused|  \\
VideoCLIP-XL \cite{wang-etal-2024-videoclip}  & \verb|alibaba-pai/VideoCLIP-XL|  \\
SigLIP 2 \cite{siglip2} & \verb|timm/ViT-L-16-SigLIP2-384|  \\
Cacophony \cite{cacophony}  &  Stage 2 checkpoint   \\
ViCLIP \cite{wang2024internvid}  & ViCLIP-B-16, InternVid-10M-FLT checkpoint   \\ \bottomrule
\end{tabular}
\end{table}

\subsection{Compute resources} \label{sec:comp_resources}

All training processes are done on a single 48 GB NVIDIA RTX-A6000 or an 80 GB A100 GPU with 4 CPUs. For Llama 3.1 8B Instruct, projector pre-training takes approximately 16 hours, and hypernetwork training takes around 4 days. For Llama 3.2 1B Instruct, projector pre-training takes approximately 7 hours, and hypernetwork training takes around 11 hours. The runtimes of methods are shown in \cref{tab:runtimes}. Although the runtime of the hypernetwork seems large, this is due to the pre-processing included in the hypernetwork data loader, e.g., processing hypernetwork embeddings, interleaving text and modality embeddings, etc.\ and not due to more FLOPS. Therefore, this additional time can be eliminated with further code optimisation, effectively diminishing the discrepancy between our method and \textit{FT Projector} and \textit{Projector} baselines. Additionally, adapter generation takes an insignificant amount of time, approximately 80 milliseconds. Feature extraction and training use \texttt{float32} precision, except for LLMs which use \texttt{bfloat16} precision.

\begin{table}[!ht]
\centering
\caption{Low-resource modality integration runtimes for full dataset sizes. The first value corresponds to Llama 3.1 8B Instruct and the second to Llama 3.2 1B Instruct.} \label{tab:runtimes}
\begin{tabular}{lrrrr}
\toprule
  & \multicolumn{1}{r}{\textbf{SydneyCaptions}} & \multicolumn{1}{r}{\textbf{CAPDELS}} & \multicolumn{1}{r}{\textbf{SensorCaps}} & \multicolumn{1}{r}{\textbf{ChEBI-20}} \\ \midrule
\textbf{Ours}         & 1h19m / 22m  & 5h23m / 57m  & 66m / 10m     & 1d8h / 7h10m     \\
\textbf{FT Projector} & 57m / 19m  & 4h49m / 49m  & 58m / 8m    & 1d2h / 5h25m      \\
\textbf{Projector} &  55m / 19m  & 4h48m / 48m   & 58m / 8m     & 1d13h / 5h26m      \\
\textbf{LoRA}      & 48m / 20m  & 4h27m / 47m   &  52m / 8m    & 1d1h / 4h33m      \\ \bottomrule
\end{tabular}
\end{table}

\subsection{Hyperparameters} \label{sec:hyparams}

The hypernetwork training and pre-trained projector hyperparameters are in Tables \ref{tab:hyparam_ft_proj_and_hyp} and \ref{tab:hyparam_pt_proj}, respectively. The unseen modality adaptation hyperparameters for full dataset sizes can be seen in Tables \ref{tab:hyparam_ft_proj_and_hyp}, \ref{tab:hyparam_proj}, and \ref{tab:hyparam_lora}. The adaptation hyperparameters are not tuned, generally following the same values except for the learning rate scheduler. As the dataset size decreases, the epochs are multiplied to keep the number of steps (approximately) constant. Better performance for smaller dataset sizes can be achieved with less training, although we do not explicitly aim for training efficiency. Additionally, the batch sizes of SydneyCaptions, CAPDELS, SensorCaps, and SoundBible setups decrease to 16 for the dataset size of 32 to allow stochasticity during adaptation.

\begin{table}[!htbp]
    \centering
    \caption{Pre-trained projector hyperparameters. When two values are provided, the first corresponds to Llama 3.1 8B Instruct and the second to Llama 3.2 1B Instruct.}
    \label{tab:hyparam_pt_proj}
    \begin{tabular*}{\linewidth}{l@{\extracolsep{\fill}}r}
    \toprule
    Optimizer & AdamW \cite{adamw} \\
    \quad ($\beta_1$, $\beta_2$) & (0.9, 0.95) \\ 
    \quad Weight decay & 5$e$-6 \\
    Learning rate & 1$e$-4 \\
    Learning rate scheduler & Linear warmup for 1k steps, then cosine decay to 0 towards the end  \\
    Warmup steps & 1000 \\
    Batch size & 32 / 64 \\
    Epochs & 5 \\
    \quad $\Rightarrow$ Steps & 108648 / 54325 \\
    Projector & \\
    \quad Architecture & 2-layer MLP with approximate GELU non-linearity \\
    \quad Hidden dimension & 768 \\
    \quad Dropout & 0.1 \\\bottomrule
    \end{tabular*}
\end{table}

\begin{table}[!htbp]
    \centering
    \caption{Hypernetwork training hyperparameters. When two values are provided, the first corresponds to Llama 3.1 8B Instruct and the second to Llama 3.2 1B Instruct.}    \label{tab:hyparam_hypnet}
    \begin{tabular*}{\linewidth}{l@{\extracolsep{\fill}}r}
    \toprule
    Optimizer &  AdamW \\
    \quad ($\beta_1$, $\beta_2$) & (0.9, 0.95) \\ 
    \quad Weight decay & 5$e$-6 \\
    Learning rate & 1$e$-4 \\
    Learning rate scheduler & Linear warmup for 1k steps, then cosine decay to 0 towards the end  \\
    Warmup steps & 1000 \\
    Batch size & 2 / 4 \\
    Gradient accumulation steps & 80 / 40 \\
    Subset batch size & 128 \\
    Epochs & 5 \\
    \quad $\Rightarrow$ Steps & 195505 / 97952 \\
    Hypernetwork & \\
    \quad Backbone & Self-attention \\
    \quad Num. heads & 1 \\
    \quad Context length & 259 (2 + 1 + 128 $\times$ 2) \\
    \quad Hidden dimension & 768 \\
    \quad Dropout & 0.1 \\ 
    \quad Generated adapter \\
    \qquad Rank & 32 \\
    \qquad Alpha & 32 \\\bottomrule
    \end{tabular*}
\end{table}

\begin{table}[!htbp]
    \centering
    \caption{\textit{FT Projector} baseline and hypernetwork adaptation hyperparameters for full dataset sizes. When two values are provided, the first corresponds to Llama 3.1 8B Instruct and the second to Llama 3.2 1B Instruct.}
    \label{tab:hyparam_ft_proj_and_hyp}
    \begin{tabular*}{\linewidth}{l@{\extracolsep{\fill}}r}
    \toprule
    Optimizer & AdamW \\
    \quad ($\beta_1$, $\beta_2$) & (0.9, 0.999) \\ 
    \quad Weight decay & 5$e$-6 \\
    Learning rate & 1$e$-4 \\
    Learning rate scheduler & Constant \\
    Dataset-specific hyperparameters & \\
    \quad SydneyCaptions \\
    \qquad Batch size & 16 / 64 \\
    \qquad Gradient accumulation steps & 1 \\
    \qquad Epochs & 60 \\
    \quad CAPDELS \\
    \qquad Batch size & 8 / 32 \\
    \qquad Gradient accumulation steps & 1 \\
    \qquad Epochs & 40 \\
    \quad SensorCaps \\
    \qquad Batch size & 8 / 32 \\
    \qquad Gradient accumulation steps & 1 \\
    \qquad Epochs & 20 \\
    \quad ChEBI-20 \\
    \qquad Batch size & 2 / 8 \\
    \qquad Gradient accumulation steps & 32 / 8 \\
    \qquad Epochs & 10 \\
    \quad SoundBible \\
    \qquad Batch size & 16 / 64 \\
    \qquad Gradient accumulation steps & 1 \\
    \qquad Epochs & 70 \\
    Projector & \\
    \quad Architecture & 2-layer MLP with approximate GELU non-linearity \\
    \quad Hidden dimension & 768 (or equal to enc. dim. if enc. dim. is smaller)  \\
    \quad Dropout & 0.1 \\\bottomrule
    \end{tabular*}
\end{table}

\begin{table}[!htbp]
    \centering
    \caption{\textit{Projector} baseline hyperparameters for full dataset sizes. When two values are provided, the first corresponds to Llama 3.1 8B Instruct and the second to Llama 3.2 1B Instruct.}
    \label{tab:hyparam_proj}
    \begin{tabular*}{\linewidth}{l@{\extracolsep{\fill}}r}
    \toprule
    Optimizer &  AdamW \\
    \quad ($\beta_1$, $\beta_2$) & (0.9, 0.999) \\ 
    \quad Weight decay & 5$e$-6 \\
    Learning rate & 1$e$-4 \\
    Learning rate scheduler & Linear warmup, then cosine decay to 0 towards the end  \\
    Dataset-specific hyperparameters & \\
    \quad SydneyCaptions \\
    \qquad Batch size & 16 / 64 \\
    \qquad Gradient accumulation steps & 1 \\
    \qquad Epochs & 60 \\
    \qquad Warm-up steps & 100 \\
    \quad CAPDELS \\
    \qquad Batch size & 8 / 32 \\
    \qquad Gradient accumulation steps & 1 \\
    \qquad Epochs & 40 \\
    \qquad Warm-up steps & 100 \\
    \quad SensorCaps \\
    \qquad Batch size & 8 / 32 \\
    \qquad Gradient accumulation steps & 1 \\
    \qquad Epochs & 20 \\
    \quad ChEBI-20 \\
    \qquad Batch size & 2 / 8 \\
    \qquad Gradient accumulation steps & 32 / 8 \\
    \qquad Epochs & 10 \\
    \qquad Warm-up steps & 500 \\
    \quad SoundBible \\
    \qquad Batch size & 16 / 64 \\
    \qquad Gradient accumulation steps & 1 \\
    \qquad Epochs & 70 \\
    Projector & \\
    \quad Architecture & 2-layer MLP with approximate GELU non-linearity \\
    \quad Hidden dimension & Equal to enc. dim. \\
    \quad Dropout & 0.1 \\\bottomrule
    \end{tabular*}
\end{table}

\begin{table}[!htbp]
    \centering
    \caption{\textit{LoRA} baseline hyperparameters for full dataset sizes. When two values are provided, the first corresponds to Llama 3.1 8B Instruct and the second to Llama 3.2 1B Instruct.}
    \label{tab:hyparam_lora}
    \begin{tabular*}{\linewidth}{l@{\extracolsep{\fill}}r}
    \toprule
    Optimizer &  AdamW \\
    \quad ($\beta_1$, $\beta_2$) & (0.9, 0.999) \\ 
    \quad Weight decay & 5$e$-6 \\
    Learning rate & 1$e$-4 \\
    Learning rate scheduler & Constant  \\
    \quad SydneyCaptions \\
    \qquad Batch size & 16 / 64 \\
    \qquad Gradient accumulation steps & 1 \\
    \qquad Epochs & 60 \\
    \quad CAPDELS \\
    \qquad Batch size & 8 / 32 \\
    \qquad Gradient accumulation steps & 1 \\
    \qquad Epochs & 40 \\
    \quad SensorCaps \\
    \qquad Batch size & 8 / 32 \\
    \qquad Gradient accumulation steps & 1 \\
    \qquad Epochs & 20 \\
    \quad ChEBI-20 \\
    \qquad Batch size & 2 / 8 \\
    \qquad Gradient accumulation steps & 32 / 8 \\
    \qquad Epochs & 10 \\
    \quad SoundBible \\
    \qquad Batch size & 16 / 64 \\
    \qquad Gradient accumulation steps & 1 \\
    \qquad Epochs & 70 \\
    Projector & \\
    \quad Architecture & 2-layer MLP with approximate GELU non-linearity \\
    \quad Hidden dimension & 768 (or equal to enc. dim. if enc. dim. is smaller)  \\
    \quad Dropout & 0.1 \\
    LoRA & \\
    \quad Rank & 32 \\
    \quad Alpha & 32 \\\bottomrule
    \end{tabular*}
\end{table}

\end{document}